\documentclass[manuscript]{acmart}

\usepackage{multirow}
\usepackage{graphicx}
\usepackage{tabularx}
\usepackage{float}
\usepackage{url}

\AtBeginDocument{%
  }

\setcopyright{acmlicensed}
\copyrightyear{2026}
\acmYear{2026}
\acmDOI{XXXXXXX.XXXXXXX}

\begin{document}

\title[Understanding the Effects of Miscalibrated AI Confidence]{Understanding the Effects of Miscalibrated AI Confidence on User Trust, Reliance, and Decision Efficacy}
\author{Jingshu Li}
\authornote{Both authors contributed equally to this research.}
\email{jingshu@u.nus.edu}
\orcid{1234-5678-9012}
\author{Yitian Yang}
\authornotemark[1]
\email{yang.yitian@u.nus.edu}
\affiliation{%
  \institution{National University of Singapore}
  \country{Singapore}
}

\author{Renwen Zhang}
\affiliation{%
  \institution{Nanyang Technological University}
  \country{Singapore}
  }
\email{renwen.zhang@ntu.edu.sg}

\author{Q. Vera Liao}
\affiliation{%
  \institution{Microsoft Research}
  \city{Ann Arbor}
  \state{Michigan}
  \country{United States}}
\email{veraliao@umich.edu}

\author{Tianqi Song}
\email{tianqi_song@u.nus.edu}
\author{Zhengtao Xu}
\email{xuzhengtao@u.nus.edu}
\affiliation{%
  \institution{National University of Singapore}
  \country{Singapore}
}

\author{Yi-chieh Lee}
\authornote{Corresponding Author}
\affiliation{%
  \institution{National University of Singapore}
  \country{Singapore}
  }
\email{yclee@nus.edu.sg}

\renewcommand{\shortauthors}{Li et al.}

\begin{abstract}
Providing well-calibrated AI confidence can help promote users' appropriate trust in and reliance on AI, which are essential for AI-assisted decision-making. However, calibrating AI confidence---providing confidence score that accurately reflects the true likelihood of AI being correct---is known to be challenging. To understand the effects of AI confidence miscalibration, we conducted our first experiment. The results indicate that miscalibrated AI confidence impairs users’ appropriate reliance and reduces AI-assisted decision-making efficacy, and AI miscalibration is difficult for users to detect. Then, in our second experiment, we examined whether communicating AI confidence calibration levels could mitigate the above issues. We find that it helps users to detect AI miscalibration. Nevertheless, since such communication decreases users' trust in uncalibrated AI, leading to high under-reliance, it does not improve the decision efficacy. We discuss design implications based on these findings and future directions to address risks and ethical concerns associated with AI miscalibration.

\end{abstract}






\maketitle
\section{Introduction}
\label{introduction}

Artificial intelligence (AI) can support human decision-makers by extracting insights from data. It is advising on and assisting with a growing range of decision-making tasks, ranging from mundane ones like outfit choices \cite{guan2016apparel, kotouza2020towards} to high-stakes ones like the selection of medical treatments \cite{loftus2020artificial, jussupow2021augmenting, bjerring2021artificial, kiani2020impact} and financial investments \cite{shanmuganathan2020behavioural, mhlanga2020industry}.
Under the paradigm known as {\it AI-assisted decision-making}, AI provides advice to human decision-makers who then make the final decisions  \cite{zhang2020effect, wang2022effects}.
A major challenge to AI-assisted decision-making is ensuring that human trust in and reliance on AI remain at appropriate levels \cite{lee2004trust, muir1987trust, zhang2020effect, humer2024reassuring}. 
Human decision-makers need to know when to rely on or disregard AI models---also referred to as the challenge of understanding the uncertainties or error boundaries of AI \cite{bansal2019updates, li2025confidence}. Uncertainty expression of AI can take various forms depending on the model. In this work, we focus on \textit{confidence score}, which estimates the probability of {\bf classification models} making a correct prediction (e.g., the model is 70\% confident about the predicted label) \cite{guo2017calibration,zhang2020effect}.
Recent studies suggest that presenting AI confidence score is one approach to facilitate appropriate user trust and reliance \cite{zhang2020effect, rechkemmer2022confidence}, so long as the confidence is well-calibrated \cite{ghosh2021uncertainty, lai2021towards}. 
 
{\it Confidence calibration} refers to the degree to which confidence score (prior judgment) corresponds to accuracy (posterior evidence), or the true likelihood of model being correct\cite{guo2017calibration}. However, achieving well-calibrated AI confidence is technically challenging, as many ML algorithms, especially deep-leaning models, are known to provide miscalibrated confidence scores \cite{jiang2021know, guo2017calibration}. In fact, developing algorithms for calibrating AI confidence has long been a focus of AI research\cite{jiang2021know, guo2017calibration,xiong2023llms}. 
However, despite these research efforts \cite{jiang2021know, guo2017calibration}, uncalibrated confidence is still pervasive in deployed AI systems, which may exhibit overconfidence (overestimating their correctness likelihood) or underconfidence (underestimating their correctness likelihood) \cite{guo2017calibration, zhang2020mix, kumar2018trainable, wang2021confident}. 

The danger of miscalibration is that human-decision maker may not be aware of the issue and take the stated confidence score as accurate, leading to inappropriate trust and reliance. However, this hypothesis has not been tested in prior work. Therefore, we conduct EXP1 ($N=126$), where participants were asked to do decision-making tasks with AI collaborators of different calibration levels. 
We found that most participants did not recognize AI calibration levels well: indeed, when collaborating with uncalibrated AI, they tended not to detect its miscalibration. This led them to over-rely on overconfident AI and to under-rely on underconfident AI, ultimately reducing the efficacy of AI-assisted decision-making.

Given users' lack of awareness on confidence miscalibration as shown in EXP1, a possible remedy is for the AI system to transparently communicate its miscalibration. We note that confidence miscalibration {\bf at the model level} can be detected by calibration metrics \cite{guo2017calibration}, while correcting or recalibrating confidence for {\bf each prediction} is often more difficult or infeasible depending on the algorithm. To explore the effectiveness of such as remedy, we conduct EXP2 ($N=126$) to investigate how communicating model calibration level helps human decision-makers to detect AI's miscalibration and affects human decision-makers' trust in and reliance on AI. We found such communication helped participants to make more sense of AI's miscalibration. But it reduced participants trust in uncalibrated AI, resulting in higher under-reliance levels on both overconfident and underconfident AI. Thus, the efficacy of AI-assisted decision-making in were not improved compared to EXP1.

Building on our results, we offer constructive suggestions for the design of future AI-assisted decision-making and discuss the risks and ethical issues that miscalibration can pose.
Our study makes the following key contributions to the field of AI-assisted decision-making and the HCI community:

\begin{itemize}
    \item We offer empirical analysis of human decision-makers’ understanding of AI confidence calibration and of AI confidence calibration’s effects on appropriate user trust and reliance, and the efficacy of AI-assisted decision-making.
    \item We show the effectiveness and drawbacks of presenting human decision-makers with information about AI confidence calibration.
    \item We provide important theoretical and design implications for improving user understanding of AI confidence calibration and forming appropriate reliance with AI.
\end{itemize}

\section{Related Work}
\subsection{AI Confidence Score as a Form of Uncertainty Estimation}

In AI-assisted decision making, AI provides information and advice to humans, who then consider this input to make the final decision \cite{ma2024you,lu2024does,jain2023effective,chiang2023two}. The motivation for having humans as the ultimate decision-makers primarily stems from concerns about ethics and responsibility, especially in high-risk decision scenarios \cite{zhang2020effect,ma2023should}. For instance, AI may offer investment advice to investors, who then integrate this guidance with their own experience to make the final decisions \cite{cao2022ai}.
The key to success in AI-assisted decision making lies in the human's correct understanding of the model's uncertainties and error boundaries: Decision-makers need to know when to rely on or disregard the model's advice, maintaining an appropriate trust in and reliance on AI \cite{bansal2019updates,zhang2020effect,lai2021towards,lee1994trust}. 
Referring to previous studies, users' trust is defined as an attitude towards AI \cite{lee1994trust,lee2004trust,muir1987trust,rechkemmer2022confidence}, while their reliance is a kind of behavior reflecting using or not using \cite{dzindolet2003role,ma2024you,lai2021towards,ma2023should}.
Inappropriate trust and reliance can compromise the efficacy of AI-assisted decisions: studies have found that human-AI teams sometimes do not outperform AI alone because humans either follow incorrect AI advice or ignore correct AI advice \cite{buccinca2021trust, bansal2021does, buccinca2020proxy, green2019principles, Bu_inca_2021}.

Past research indicates that displaying reliable uncertainty estimations of AI’s predictions can help humans recognize its error boundaries and thus develop appropriate trust in and reliance on it \cite{ghosh2021uncertainty,lai2021towards,wischnewski2023measuring,ma2024you,rastogi2022deciding}. In AI-assisted decision making, the most common form of AI uncertainty estimation is the confidence score of classification models \cite{lai2021towards,bansal2021does,buccinca2021trust,arshad2015investigating,lee2021human}, which is also the form of AI uncertainty estimation utilized in this study. 
Decision-making AI based on classification models can output the conditional probability of a single prediction as its confidence score, reflecting the estimation of the likelihood that the prediction is correct \cite{guo2017calibration,zhang2020effect}.

To foster human decision-makers' appropriate trust in and reliance on AI, one ideal optimization is that individuals rely on AI when AI confidence score exceeds individuals' own correctness likelihood, and use their own judgment when AI confidence score is lower \cite{ma2023should,zhang2020effect,guo2017calibration}. 
Previous studies have reported that when AI’s confidence is relatively high, its users are more likely both to believe in the correctness of that advice \cite{rechkemmer2022confidence} and accept such advice \cite{zhang2020effect}.
Various prior studies of AI-assisted decision making have presented AI confidence scores to users in either experimental or practical settings to help users capture the uncertainty of AI \cite{lai2021towards,bansal2021does} and explored various visualization techniques for making such levels more understandable \cite{zhao2023evaluating}.
However, most previous research of this kind has been built on the ideal assumption that AI confidence is well-calibrated \cite{lai2021towards,ghosh2021uncertainty}.

Unlike the ideal assumptions in previous research, achieving well-calibrated AI confidence scores in practice is a challenging technical problem \cite{jiang2021know, guo2017calibration}. AI models can be uncalibrated: some models exhibit overconfidence, overestimating their probability to make correct predictions \cite{guo2017calibration, zhang2020mix, kumar2018trainable}; some are underconfident, underestimating their correctness likelihood \cite{wang2021confident}. 
Research indicates that the increase in model capacity, along with lack of regularization, is closely related to the problem of miscalibration \cite{guo2017calibration}.
Addressing the miscalibration issue of AI models is a common focus in the fields of machine learning and AI \cite{jiang2021know, guo2017calibration, xiong2023llms}.
These uncalibrated AI confidence scores, or unreliable expressions of AI uncertainty, can be hazardous \cite{ma2024you}. However, to date, it is unclear whether human decision-makers can detect the miscalibration of AI confidence score, nor is there a deep understanding of how uncalibrated AI confidence scores hinder AI-assisted decision making. Therefore, this study asks:

\begin{itemize}
    \item {\bf RQ1}: To what extend can users detect AI confidence miscalibration?
    \item {\bf RQ2}: How does AI confidence miscalibration affect users' (a) trust in and (b) reliance on AI, and (c) the efficacy of AI-assisted decision making?
\end{itemize}

\subsection{Improving Users' Understanding of Models through Communicating Information about Models}
Some previous research has also shown that in addition to displaying an AI model's uncertainty of predictions, communicating relevant information about the AI model with human decision-makers can help them better understand the AI model, calibrate their trust in and reliance on AI, and achieve more effective AI-assisted decision making \cite{lai2021towards,rechkemmer2022confidence}. Such information includes the model's performance \cite{rechkemmer2022confidence,yin2019understanding,lai2019human,harrison2020empirical,lai2020chicago}, features and structure \cite{slack2019assessing,lakkaraju2016interpretable,poursabzi2021manipulating}, the training data used by the model \cite{dodge2019explaining,kulesza2013too}, and so on. For example, it has been reported that the stated accuracy of an AI model---that is, the accuracy performance indicator directly presented to human decision-makers before collaboration---significantly influences humans' trust in and reliance on AI (i.e., the adoption of AI predictions) \cite{rechkemmer2022confidence,yin2019understanding,lai2019human}. The higher the stated accuracy of the AI, the greater the level of trust and reliance human decision-makers have on the AI \cite{rechkemmer2022confidence,yin2019understanding}.

At the model level, AI confidence calibration level can be detected by calibration metrics \cite{guo2017calibration}, while correcting or recalibrating confidence for each prediction is often more difficult or infeasible depending on the algorithm. Given users' lack of awareness on confidence miscalibration as we found, communicating AI confidence calibration levels to human decision-makers may help users detect AI confidence miscalibration and its potential issues. This, in turn, could enable users to make reasonable use of possibly uncalibrated AI confidence, reducing the harms associated with it, thereby fostering a appropriate trust in and reliance on AI, and achieving better efficacy in AI-assisted decision making. Consequently, we propose:

\begin{itemize}
    \item {\bf RQ3}: How does communicating AI confidence calibration levels affect users’ understanding of such calibration?
    \item {\bf RQ4}: How does communicating AI confidence calibration levels influence users' (a) trust in and (b) reliance on AI, and (c) the efficacy of AI-assisted decision making?
    \item {\bf RQ5}: With communication about AI confidence calibration levels, how does AI confidence miscalibration affect users' (a) trust in and (b) reliance on AI, and (c) the efficacy of AI-assisted decision making?
\end{itemize}

\section{Experiment 1: Users' Awareness of AI Confidence Miscalibration and Effects of Uncalibrated AI Confidence Score}
The purpose of our first experiment was to address {\it RQ1} and {\it RQ2}, as explained below.

\subsection{Experimental Design}
\subsubsection{Participants}
Participants of EXP1 were recruited online from the Prolific platform, meeting the following criteria: (1) aged between 21 and 60 years, (2) English as their first language, and (3) able to use a personal computer. We ensured gender balance during the recruitment. Each participant could take part in the experiment only once. The experiment was expected to last 25 minutes, with participants compensated at Prolific's recommended rate of £9 per hour. After excluding participants who did not complete the task for various reasons and those who failed multiple attention checks, EXP1 comprised 126 unique participants (42 per condition). Of these, 53.2\% were female, with an average age of 32.4 years (SD=10.0), and 72.2\% held at least a bachelor's degree.

\subsubsection{Experimental Task}
Inspired by previous research~\cite{vodrahalli2022uncalibrated,vodrahalli2022humans,lai2021towards}, the city image recognition task was employed as the task for AI-assisted decision making in this study.
Participants were asked to recognize which of three U.S. cities---New York, Chicago, or San Francisco---a given image depicted.
Images were sourced from one of these cities and included urban architectural or geographical features and were obtained from publicly available datasets and online resources~\cite{vodrahalli2022humans}.
The task offers the advantage of requiring minimal domain-specific expertise, allowing participants with basic training to competently perform and making it suitable for online randomized behavioral experiments. Without AI assistance, the average accuracy of participants on city image recognition was 65.1\%.

\subsubsection{Procedure}
\begin{figure}[t]
\centering 
\includegraphics[width=0.65\textwidth]{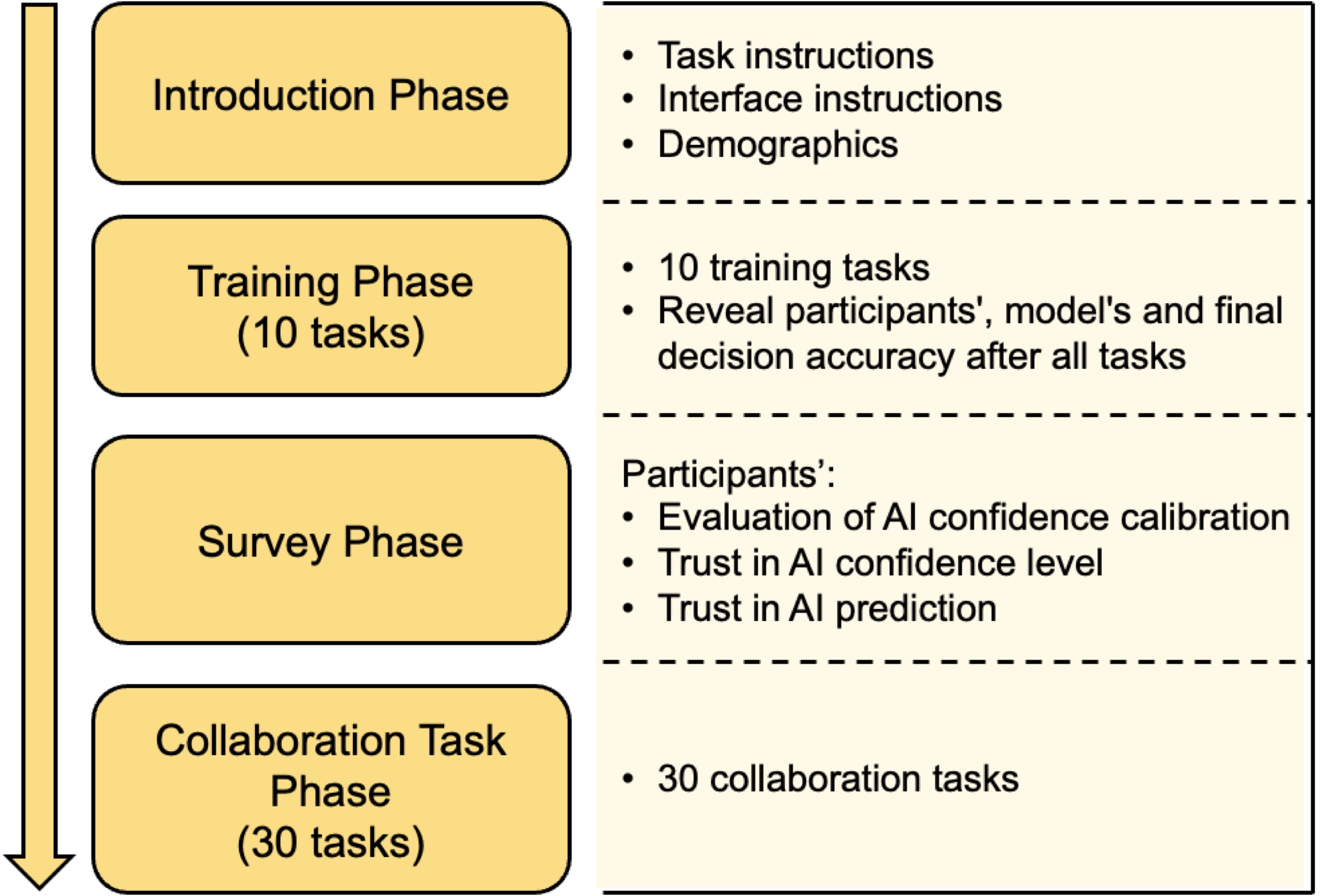} 
\caption{Flowchart illustrating the experimental procedure. }
\label{procedure}
\Description{This figure is a flowchart illustrating the experimental procedure, from upper to bottom, there are 4 phases of the experimental procedure. The first phase is the Introduction phase, including task instructions, interface instructions for participants and collecting participants' demographics. The second phase is the Training Phase, including 10 training tasks and performance feedback after all tasks. The third phase is the survey phase, including a survey which asks participants to report their evaluation of AI confidence calibration levels, their trust in AI confidence score and in AI prediction. The last phase is the Collaboration Task Phase, including 30 collaboration tasks.}
\end{figure}

As shown in Fig.~\ref{procedure}, EXP1 was organized into four sequential phases: Introduction Phase, Training Phase, Survey Phase, and Collaboration Task Phase. Upon accepting the task, participants were first asked to sign the consent form and then they were redirected into our experimental website to start the Introduction phase.

\textbf{Introduction Phase:}
In this phase, participants were initially provided with written explanations about the objective of the prediction task, which was to identify the city from which a given image originated. We introduced the three U.S. cities (New York, Chicago, and San Francisco) serving as the sources of the images, highlighting their distinctive landscapes and geographical features, along with examples of the images. 
Then the concepts about AI confidence calibration levels were introduced and the corresponding terms were explained as below:

\begin{itemize}
    \item When the AI's average confidence score is lower than the accuracy of its predictions, we term it 'underconfident.'
    \item When the AI's average confidence score aligns closely with the accuracy of its predictions, we describe it as 'well-calibrated.'
    \item When the AI's average confidence score exceeds the accuracy of its predictions, we label this 'overconfident.' 
\end{itemize}

Participants were also given written instructions on how to use our experimental interface and collaborate with AI to complete the decision-making task. Before moving to the next phase, participants were required to correctly answer three fact-checking questions about the instructions to ensure they had attentively learned and understood the material. Demographic information of the participants was also collected during this stage.

\textbf{Training Phase:}
In this phase, participants were given 10 training tasks to practice.
Following the paradigm of AI-assisted decision making from previous studies \cite{ma2023should,ma2024you,zhang2020effect,rechkemmer2022confidence,lai2021towards}, participants were required to make an initial decision independently based on the provided image for each task. Subsequently, they were presented AI's prediction and its confidence score. Finally, participants were asked to make a final decision based on the AI's advice and their initial decision.
After completing all the training tasks, according to previous research, participants were informed about their accuracy and the AI model's accuracy over these 10 training tasks \cite{zhang2020effect,ma2024you,rechkemmer2022confidence}. During this process, participants could assess the AI model's performance and adjust their trust and reliance on it based on the model's accuracy and their own consistency with the model \cite{zhang2020effect,pescetelli2021role}.

\textbf{Survey Phase:}
Following the Training Phase, participants were asked to complete a survey designed to assess their detection of AI confidence miscalibration and capture their trust (attitude) towards their AI collaborator. 
In the survey, participants were to report their evaluation of the AI collaborator's confidence calibration level, that is, whether they perceived their AI collaborator to be underconfident, well-calibrated, or overconfident. 
Meanwhile, they were required to report their trust in the accuracy of the AI's predictions and their trust in the confidence scores expressed by the AI.

\textbf{Collaboration Task Phase:}
Finally, participants proceeded to complete 30 city image recognition tasks. This phase aimed to observe participants' reliance on AI and the efficacy of AI-assisted decision making. Participants were told they would collaborate with the same AI as the one in Training Phase. The decision-making process was: for each task, participants first made their own decision, then were presented with the AI's advice and confidence score, and subsequently made their final decision. Upon completing all tasks, the experiment concluded, and participants exited the experiment system.

\subsubsection{Experimental Interface}
\begin{figure}[t]
\centering 
\includegraphics[width=0.9\textwidth]{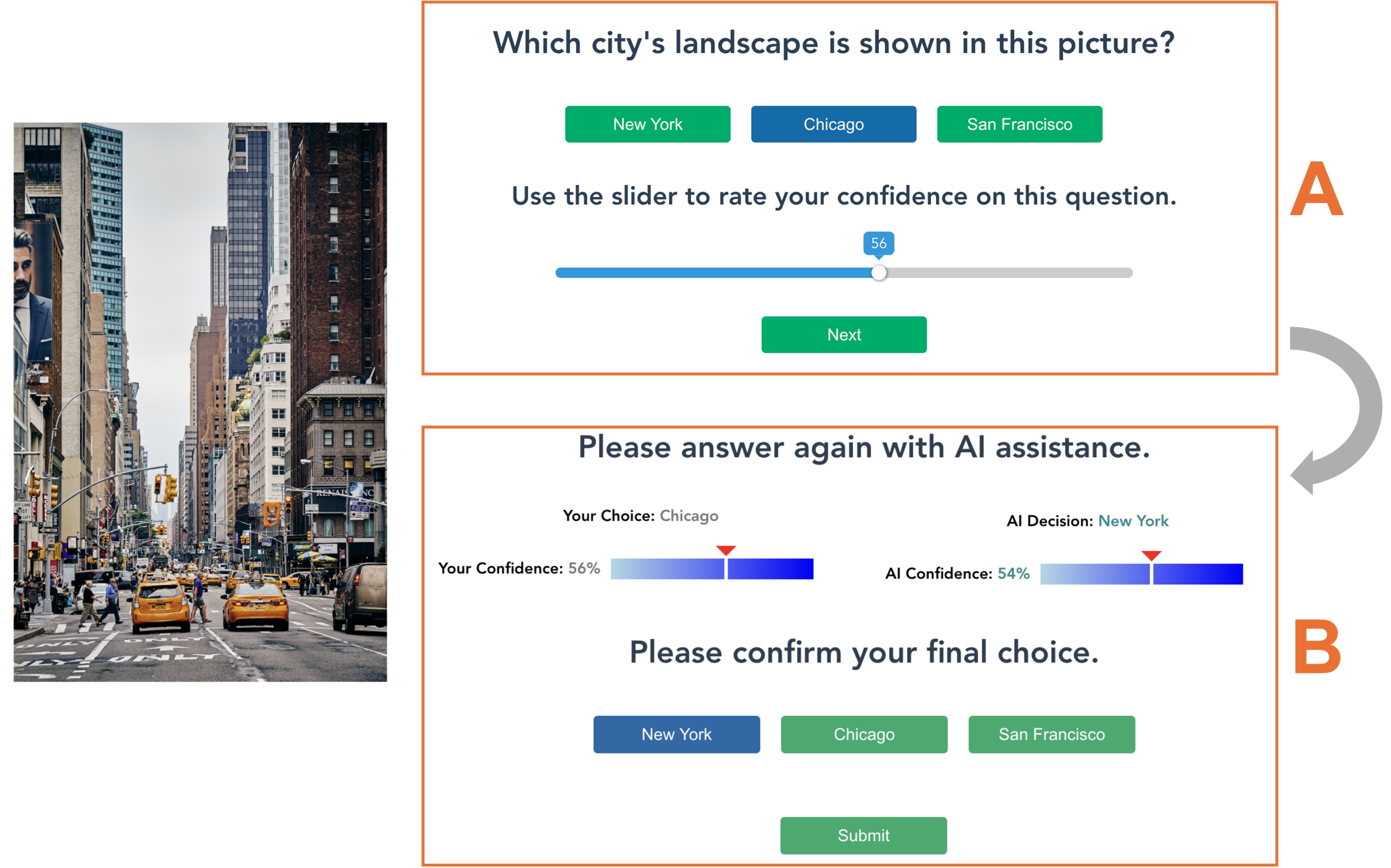} 
\caption{Interface of the city image recognition task (an example on a task instance). The image to recognize is shown on the left. {\bf A}: Participants were asked to make their first decision here once the city image was displayed. They should make a decision using the green buttons and then indicate their confidence in their decision using the slider. {\bf B}: After clicking the Next button in A, participants were asked to make the final decision according to their first decision and AI's suggestion with confidence. They should make the final decision using the green buttons and then click the submit to finish the task.}
\label{interface}
\Description{Interface of the city image recognition task. The image to recognize is shown on the left. In part A, participants were asked to make their first decision here once the city image was displayed. They should make a decision using the green buttons and then indicate their confidence in their decision using the slider. In part B, after clicking the Next button in A, participants were asked to make the final decision according to their first decision and AI's suggestion with confidence. They should make the final decision using the green buttons and then click the submit to finish the task.}
\end{figure}

To support the experiment, we implemented an online experimental system using the {\it JavaScript} framework {\it Vue.js}, with the web interface depicted in Figure \ref{interface}. For each task, at first, the system displayed the task image and associated choices to participants and participants were asked to make their first decision. Inspired by previous research \cite{buccinca2021trust, ma2023should, ma2024you, chong2022human}, participants were also asked to report their confidence in the first decision through a slider to promote careful thought and cautious decision-making, and to serve as a reference when making their final decision. Next, the system presented AI suggestion and confidence score, and then asked participants to make a final decision. This system captured both the first and final decisions of participants and manages backend storage using {\it MySQL}. Additionally, informed consent and surveys were conducted on the Qualtrics~\footnote{\url{https://www.qualtrics.com/}} online survey platform.

\subsubsection{Experimental Conditions and AI Model}
Drawing on methodologies from previous research \cite{buccinca2021trust,bussone2015role,lai2021towards}, this study employed a {\it simulated AI} and manually generated predictions and confidence scores for the AI model. This approach facilitates the control of the AI model's accuracy and confidence scores, thereby manipulating AI confidence calibration levels.
The AI's accuracy remained constant at 70\% during both the Training Phase and the Collaboration Task Phase.
There were three different conditions of AI confidence calibration levels for the between-subject design: \textbf{underconfident} (EXP1-U), \textbf{well-calibrated} (EXP1-W), and \textbf{overconfident} (EXP1-O). According to definitions and research on AI confidence calibration \cite{guo2017calibration,wang2021confident}, the definitions for these conditions in our experiment depend on the relationship between the AI's average confidence score and its accuracy.

\begin{itemize}
    \item \textbf{Underconfident AI:} An underconfident AI was defined as having an average confidence score below its accuracy rate. Specifically, under this condition, the AI's average confidence score during both the Training Phase and Collaboration Task Phase was 60\%, which was below its accuracy.
    \item \textbf{Well-calibrated AI:} A well-calibrated AI was defined as having an average confidence score that exactly matches its accuracy. Specifically, under this condition, the AI's average confidence score during both the Training Phase and Collaboration Task Phase was 70\%, equal to its accuracy.
    \item \textbf{Overconfident AI:} An overconfident AI was defined as having an average confidence score above its accuracy. Specifically, under this condition, the AI's average confidence score during both the Training Phase and Collaboration Task Phase was 80\%, greater than its accuracy.
\end{itemize}

Note that in this setup, the AI's \textbf{average} confidence score was manipulated. Rather than maintaining the confidence score for each task at the average for that condition, the confidence scores of the AI could vary among different tasks. For each condition, the standard deviation of the AI's confidence score was 7.770\%, with a range of $\pm$15\% around the average for that condition.

\subsubsection{Measurements}
To address RQ1 and RQ2, this study measures participants' detection of AI confidence miscalibration and their trust in AI through a questionnaire. Additionally, participants' reliance on AI and the efficacy of AI-assisted decision making are assessed based on their behavior during the Collaboration Task Phase. Further details are provided below:

\paragraph{Participants' Detection of AI Confidence Misalibration} In the Survey Phase, participants were asked to evaluate the confidence calibration levels of the AI with which they interacted. They were provided with three options: underconfident, well-calibrated, and overconfident. Necessary explanations about these three terms were also provided according to their definitions (e.g. "When AI's mean confidence score is higher than its accuracy of its prediction, we call this 'overconfident.' ... For example, AI might claim to be 90\% confident about its prediction, but in reality, it is correct only 70\% of the time." See details in Appendix~\ref{app: survey}). Participants must select only one of these options.  

\paragraph{Participants' Trust in AI} Drawing from theories in past studies, participants' trust in AI, as an attitude \cite{mcknight2002developing, madsen2000measuring, lee1994trust, lee2004trust, muir1987trust}, can be categorized into trust in different aspects of AI's capabilities. Given that this study both presented AI's predictions and AI confidence scores, participants' trust in AI had been divided into \textbf{trust in AI predictions} and \textbf{trust in AI confidence scores}, measured separately. 

Trust in AI predictions refers to how much trust participants have in the correctness of the AI predictions. A 3-item 7-point Likert scale (1="Strongly disagree",7="Strongly agree"; $M=4.869$, $SD=0.856$; McDonald's $\omega=0.796$, 95\% CI: [0.754, 0.839]) was employed to measure it, which was adapted from questions and scales previously utilized in measuring human trust in AI or automation \cite{mcknight2002developing, madsen2000measuring, yin2019understanding}.
Trust in AI confidence scores assesses how much trust participants have in presented AI confidence scores. Similarly, a 3-item 7-point Likert scale (1="Strongly disagree",7="Strongly agree"; $M=4.718$, $SD=1.018$; McDonald's $\omega=0.862$, 95\% CI: [0.833, 0.892]) adapted from previous studies was employed to measure it \cite{mcknight2002developing, madsen2000measuring, yin2019understanding}. Details of these scales are provided in the Appendix~\ref{app: survey}.

\paragraph{Participants' Reliance on AI} 
Refer to prior research, this experiment utilized \textbf{switch percentage} as an indicator to measure participants' reliance on AI \cite{zhang2020effect,ma2024you}. The switch percentage refers to the percentage of tasks in which participants switch their final decision to AI advice when their first decisions disagree with AI advice, with higher values indicating greater reliance on AI. Previous studies have also used the agreement percentage, which measures the percentage of tasks in which participants' final decisions agree with AI advice \cite{zhang2020effect,ma2024you}. However, previous research suggests that switch percentage is a stricter measure than agreement percentage, as the former only accounts for cases where participants actively change their decisions, whereas the latter includes instances where the first decision already coincides with the AI suggestion \cite{zhang2020effect}. Therefore, this research exclusively employed switch percentage as the metric.

\begin{align}
    \text{Switch Percentage} &=
    \frac{\parbox{0.6\textwidth}{\fontsize{7.8}{9.8}\selectfont \centering \textit{\# of Tasks the Participant Switched Final Decision to AI Advice}}}
         {\parbox{0.6\textwidth}{\fontsize{7.8}{9.8}\selectfont \centering \textit{\# of Tasks the Participant's First Decision Disagreed with AI Advice}}}
\end{align}

Further, to identify participants' inappropriate reliance on AI, this study employed two additional metrics: \textbf{over-reliance percentage} and \textbf{under-reliance percentage}, derived from previous study \cite{ma2024you}. 
Over-reliance percentage is defined as the percentage of tasks where the participant switched to the AI's incorrect advice among all tasks where the participant switched, with higher percentages indicating over-reliance on AI. Under-reliance percentage is defined as the percentage of tasks where the participant made wrong final decision with correct AI advice among all tasks where the AI gave correct advice, with higher percentages indicating a higher degree of under-reliance on AI.

\begin{align}
    \text{Over-reliance Percentage} &= 
    \frac{\parbox{0.6\textwidth}{\fontsize{7.8}{9.8}\selectfont \centering \textit{\# of Tasks the Participant Switched Final Decision to Incorrect AI Advice}}}
         {\parbox{0.6\textwidth}{\fontsize{7.8}{9.8}\selectfont \centering \textit{\# of Tasks the Participant Switched Final Decision to AI Advice}}}
\end{align}

\begin{align}
    \text{Under-reliance Percentage} &=
    \frac{\parbox{0.6\textwidth}{\fontsize{7.8}{9.8}\selectfont \centering \textit{\# of Tasks the Participant Made Incorrect Final Decision with AI Correct Advice}}}
         {\parbox{0.6\textwidth}{\fontsize{7.8}{9.8}\selectfont \centering \textit{\# of Tasks with AI Correct Advice}}}
\end{align}

\paragraph{AI-assisted Decision Making Efficacy} 
This study assessed the efficacy of AI-assisted decision making using the \textbf{increase of accuracy after AI advice}, defined as the difference between the joint accuracy of human-AI teams and the accuracy of participants without AI assistance. This metric aligns with accuracy-based assessment methods used in many previous studies and offers an intuitive measure of performance \cite{lai2021towards}.
\begin{align}
    \text{Accuracy Increase} &= \text{\fontsize{7.8}{9.8}\selectfont \textit{Final Decision Accuracy}} - \text{\fontsize{7.8}{9.8}\selectfont \textit{First Decision Accuracy}}
\end{align}

\subsubsection{Analysis}
In statistical analysis strategy, statistal tests, including ANOVA for scale measurements and $\chi^2$ test for categorical measurements, were employed to capture the difference between treatments. 
For ANOVA's assumption check, given that each group in our study consisted of 42 participants, our sample size was sufficiently large ($>30$) for the normality assumption \cite{ghasemi2012normality}. 
To examine the homogeneity of the variances, we used Levene's test in different confidence calibration level groups for each dependent variable \cite{02462954-2341-35c6-ac32-7bc139c0d8ab}. When the assumption of equal variances was satisfied ($p$-value $>$ 0.05), we proceeded with a standard ANOVA \cite{fisher1970statistical} followed by Tukey's Honest Significant Difference (HSD) test \cite{tukey1949comparing} for post hoc comparisons. 
If the homogeneity assumption was not met, Welch's ANOVA was used \cite{welch1951comparison}, coupled with the Games-Howell \cite{games1976pairwise} post hoc test. 

Our analysis strategy also included linear regression (LR), which models the relationship between a dependent variable and independent variables. It calculates regression coefficients to understand the direction and magnitude of these relationships. The significance of these coefficients, assessed through $p$-values, determines the reliability of the observed relationships. The $p$-values are calculated based on the t-statistic, which emerges from the regression coefficient divided by the standard error of this coefficient.

\subsection{Results of EXP1}
\subsubsection{Participants' Detection of AI Confidence Misalibration (RQ1)}
\begin{figure}[t]
\centering 
\includegraphics[width=0.5\textwidth]{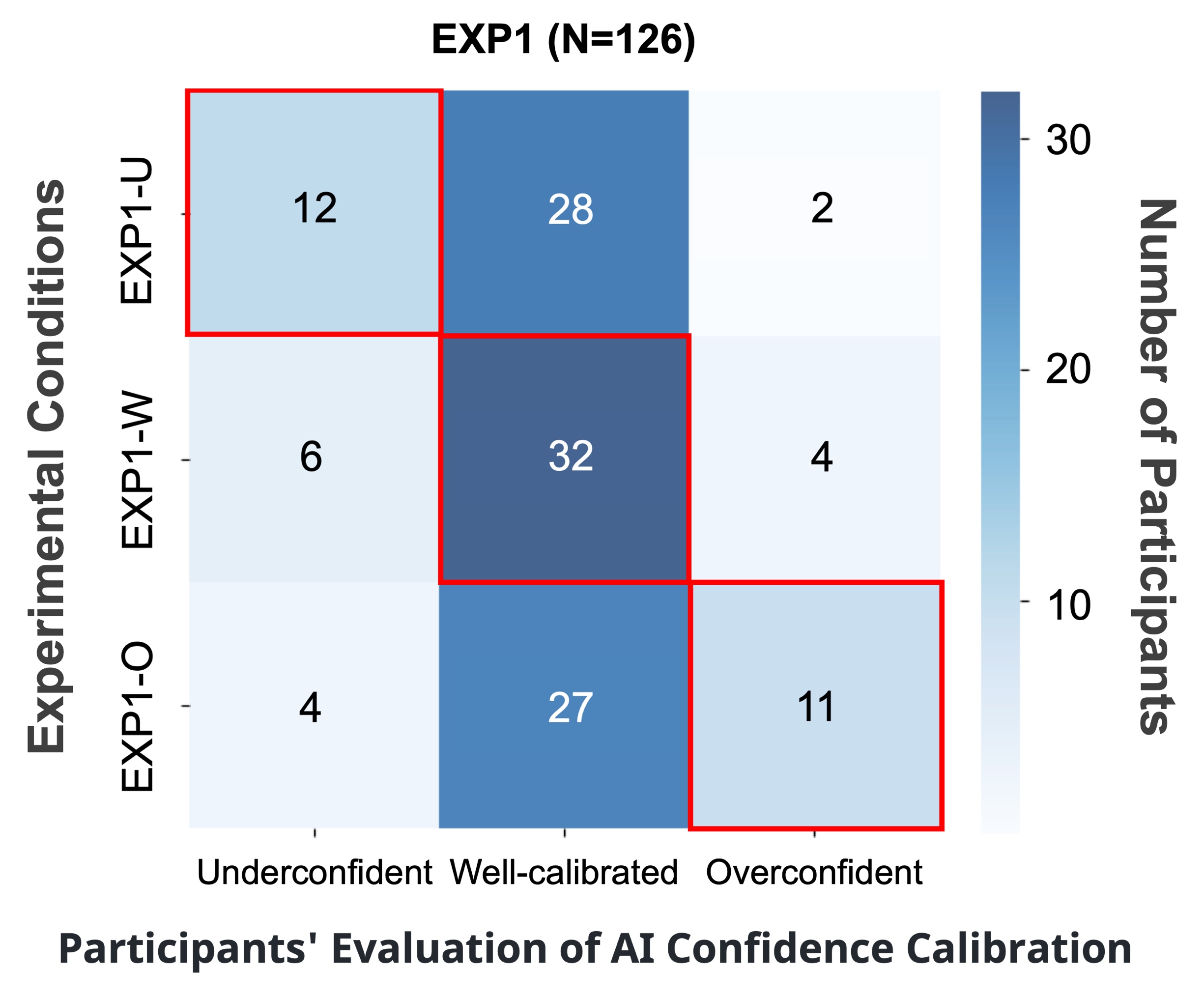} 
\caption{Participants' evaluations of AI confidence calibration level in Experiment 1, shown as a heatmap. The rows represent the three experimental conditions, and the columns indicate participants' evaluations of AI confidence calibration level. The numbers in each cell represent the count of participants giving that particular evaluation in corresponding experimental condition. The darker blue colors indicate a higher number of participants. Correct evaluations in corresponding conditions are highlighted with red boxes.}
\label{heatmap1}
\Description{This figure represents participants' evaluations of AI confidence calibration in Experiment 1, shown as a heatmap. The rows represent the three experimental conditions: the underconfident condition, the well-calirbated condition and the overconfident condition. The columns indicate participants' evaluations of AI confidence calibration: underconfident, well-calibrated, and overconfident. The numbers in each cell represent the count of participants giving that particular evaluation in corresponding experimental condition (from left to right, upper to bottom, the numbers are: 12, 28, 2, 6, 32, 4, 4, 27, 11). The darker blue colors indicate a higher number of participants. Correct evaluations in corresponding conditions are highlighted with red boxes. The underconfident condition underconfident evaluation cell, the well-calibrated condition well-calibrated evaluation cell and the overconfident condition overconfident evaluation cell are highlighted.}
\end{figure}

As shown in Figure~\ref{heatmap1}, under the condition of well-calibrated AI (i.e., in EXP1-W), 32 participants (76.2\%) reported the AI was well-calibrated, 6 participants (14.3\%) considered the AI to be underconfident, and 4 participants (9.5\%) thought the AI was overconfident. When the AI was underconfident (EXP-U), a $\chi^2$ test revealed that the distribution of evaluations on AI confidence calibration level in EXP1-U did not significantly differ from that in EXP1-W ($\chi^2(2, 84) = 2.933$, $p = 0.231$). 12 participants (28.6\%) considered the AI underconfident, 28 participants (66.7\%) incorrectly reported the AI was well-calibrated, and 2 participants (4.7\%) thought the AI was overconfident. Similar results were observed when the AI was overconfident (EXP1-O); the $\chi^2$ test indicated no significant difference in the distribution of evaluations on AI confidence calibration level between EXP1-O and EXP1-W ($\chi^2(2, 84) = 4.090$, $p = 0.129$). 11 participants (26.2\%) reported the AI was overconfident, 27 participants (64.3\%) erroneously thought the AI was well-calibrated, and 4 participants (9.5\%) considered it underconfident.

Despite the AI exhibiting overconfident or underconfident confidence scores, the majority of participants still regarded the AI as well-calibrated. \textbf{This result suggests that many participants face challenges in detecting AI confidence miscalibration, and they are struggling to identify inconsistencies between AI confidence scores and AI accuracy when AI is uncalibrated.}

\subsubsection{The Effects of AI Confidence Miscalibration on Participants' Trust in AI (RQ2.a)}
\label{sec:RQ2.a}
\begin{figure}[t]
\centering 
\includegraphics[width=0.95\textwidth]{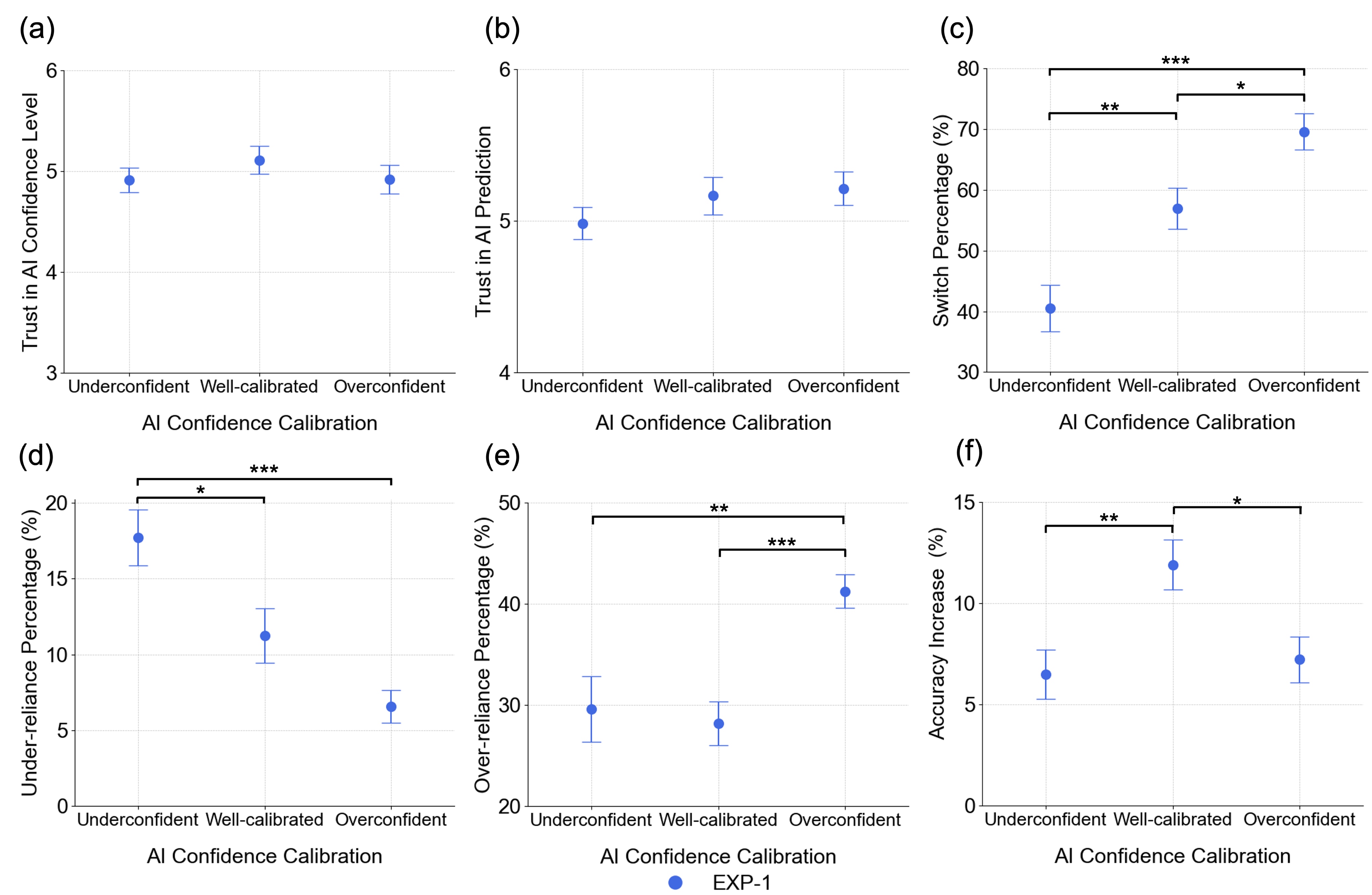} 
\caption{Experiment 1 results across three conditions. The dots represent the mean values, and the error bars show one standard error. The significance levels are labeled ($p<0.05$: *, $p<0.01$: **, $p<0.001$: ***).
{\bf (a)} Trust in AI confidence score.
{\bf (b)} Trust in AI prediction.
{\bf (c)} Switch percentage.
{\bf (d)} Under-reliance percentage.
{\bf (e)} Over-reliance percentage.
{\bf (f)} Accuracy increase.
}
\label{exp1_result}
\Description{There are six subfigures in this figure, showing the results of experiment 1, including trust in AI confidence score, trust in AI prediction, switch percentage, under-reliance percentage, over-reliance percentage, and accuracy increase, as in order. In each subfigure, the dots represent the mean value, and the error bars show one standard error. The significance levels are labeled ($p<0.05$: *, $p<0.01$: **, $p<0.001$: ***). The specific results are given in the results subsection.}
\end{figure}

The results of the one-way ANOVA indicated that, in EXP1, there was no significant difference in participants' trust in AI prediction across different conditions ($F(2, 123) = 1.127$, $p = 0.327$, $\eta^2 = 0.018$), as shown in Fig.~\ref{exp1_result} (a) and (b). Similarly, participants' trust in AI confidence scores did not differ significantly across conditions ($F(2, 123) = 0.690$, $p = 0.503$, $\eta^2 = 0.011$).
\textbf{The miscalibration of the AI did not significantly influence participants' trust in AI predictions or their trust in AI confidence scores.}

\subsubsection{The Effects of AI Confidence Miscalibration on Participants' Reliance on AI (RQ2.b)}
\label{sec:RQ2.b}
Regarding participants' reliance on AI, we initially focused on their switch percentage, as shown in Fig.~\ref{exp1_result} (c). One-way ANOVA results revealed a significant main effect of AI confidence miscalibration on participants' switch rates ($F(2, 123) = 18.354$, $p < 0.001$, $\eta^2 = 0.230$). Post-hoc analysis indicated that the switch percentage in EXP1-U ($M = 40.492\%$, $SD = 24.752\%$) was significantly lower ($t = -3.422$, $p = 0.002$) than in EXP1-W ($M = 56.975\%$, $SD = 21.868\%$). Conversely, the switch percentage in EXP1-O ($M = 69.587\%$, $SD = 19.246\%$) was significantly higher ($t = 2.619$, $p = 0.027$) than in EXP1-W ($M = 56.975\%$, $SD = 21.868\%$).
Also, the switch percentage in EXP1-O ($M = 69.587\%$, $SD = 19.246\%$) was significantly higher ($t = 6.041$, $p < 0.001$) than in EXP1-U ($M = 40.492\%$, $SD = 24.752\%$).

Regarding participants' under-reliance percentage, one-way Welch's ANOVA (not passing the Homogeneity test, Levene's test result: $F(2, 123) = 5.251$, $p = 0.006$) indicated a significant main effect of AI confidence miscalibration ($F(2, 123) = 13.886$, $p < 0.001$, $\eta^2 = 0.163$), as shown in Fig.~\ref{exp1_result} (d). Post-hoc analysis revealed that the under-reliance percentage in EXP1-U ($M = 17.687\%$, $SD = 11.912\%$) was significantly higher ($t = 2.508$, $p = 0.037$) than in EXP1-W ($M = 11.224\%$, $SD = 11.703\%$) and also significantly higher ($t = 5.217$, $p < 0.001$) than in EXP1-O ($M = 6.576\%$, $SD = 6.973\%$). No significant difference ($t=2.211$, $p=0.077$) was found between the under-reliance percentages in EXP1-W and EXP1-O.

Furthermore, concerning participants' over-reliance percentage, one-way Welch's ANOVA (not passing the Homogeneity test, Levene's test result: $F(2, 123) = 7.892$, $p < 0.001$) showed a significant main effect of AI confidence miscalibration ($F(2, 77.508) = 13.156$, $p < 0.001$, $\eta^2 = 0.122$), as shown in Fig.~\ref{exp1_result} (e). Post-hoc analysis showed that the over-reliance percentage in EXP1-O ($M = 41.257\%$, $SD = 10.835\%$) was significantly higher ($t = 4.777$, $p < 0.001$) than in EXP1-W ($M = 28.207\%$, $SD = 14.002\%$) and also significantly higher ($t = 3.194$, $p = 0.006$) than in EXP1-U ($M = 29.632\%$, $SD = 20.953\%$). No significant difference was observed between the over-reliance percentage in EXP1-U and EXP1-W ($t=0.367$, $p=0.929$).

Additionally, the results from linear regression indicated a significant negative linear correlation between under-reliance percentage and switch percentage (Pearson's $r = -0.765$, $p < 0.001$), whereas a significant positive linear correlation was observed between over-reliance percentage and switch percentage ($r = 0.531$, $p < 0.001$).

\textbf{These findings suggest that uncalibrated AI impairs pariticipants' appropriate reliance on AI. Underconfident AI decreases participants' reliance on AI and leads to increased instances of under-reliance, where participants tended not to adopt correct AI suggestions. Conversely, overconfident AI increases participants' reliance on AI and leads to heightened over-reliance, where participants tended to adopt incorrect AI suggestions.}

\subsubsection{The Effects of AI Confidence Miscalibration on AI-assisted Decision Making Efficacy (RQ2.c)}
\label{sec:RQ2.c}
One-way ANOVA results showed a significant main effect of AI confidence miscalibration on the increase of accuracy after AI advice ($F(2, 123) = 6.001$, $p = 0.003$, $\eta^2 = 0.089$), as shown in Fig.~\ref{exp1_result} (f). Post-hoc analysis indicated that the accuracy increase in EXP1-U ($M = 6.508\%$, $SD = 7.859\%$) was significantly lower ($t = -3.189$, $p = 0.005$) than in EXP1-W ($M = 11.905\%$, $SD = 7.967\%$). The accuracy increase in EXP1-O ($M = 7.211\%$, $SD = 7.433\%$) was also significantly lower ($t = -2.767$, $p = 0.018$) than in EXP1-W ($M = 11.905\%$, $SD = 7.967\%$). No significant difference was observed between the accuracy increases in EXP1-O and EXP1-U ($t=0.421$, $p=0.907$).
Additionally, the results from linear regression showed that the increase of accuracy after AI advice was significantly negatively correlated with both participants' under-reliance percentage ($r = -0.229$, $p = 0.010$) and over-reliance percentage ($r = -0.426$, $p < 0.001$).

\textbf{These findings suggest that uncalibrated AI impairs efficacy in AI-assisted decision making, which is related to participants' inappropriate reliance on AI. Both overconfident and underconfident AI diminish the enhancement of human decision accuracy provided by AI advisors.}

\section{Experiment 2: Effects of Communicating AI Confidence Calibration Levels}

The purpose of our experiment 2 (EXP2) was to address \textbf{RQ3-RQ5}, serving as an extension experiment of EXP1 with additional treatments. We aimed to explore how communicating AI confidence calibration levels with users aids their detection of AI confidence miscalibration. 
Meanwhile, we further investigated whether such communication facilitated participants' appropriate trust and reliance on AI and enhanced efficacy in AI-assisted decision making with uncalibrated AI.
Additionally, we also assessed that with such communication, the effect of AI confidence miscalibration on users' trust and reliance on AI, as well as the efficacy of AI-assisted decision making.

\subsection{Experimental Design}

The setup for EXP2 was modified based on EXP1. EXP2 also had three conditions based on AI confidence calibration levels: underconfident AI (EXP2-U), well-calibrated AI (EXP2-W), and overconfident AI (EXP2-O). 
The only difference between EXP2 and EXP1 is that, during the Introduction Phase, participants additionally received information about the confidence calibration level of the AI collaborator they would be working with, as an intervention for communicating AI confidence calibration levels~\cite{rechkemmer2022confidence}. The remaining experimental setup in EXP2 was identical to that of EXP1.

\begin{itemize}
    \item For participants collaborating with underconfident AI (in EXP2-U), they received the information: \textit{"You will collaborate with an {\bf underconfident} AI in the following tasks"}
    \item For participants working with well-calibrated AI (in EXP2-W), they were informed: \textit{"You will collaborate with a {\bf well-calibrated} AI in the following tasks"}
    \item For those engaging with overconfident AI (in EXP2-O), the message was: \textit{"You will collaborate with an {\bf overconfident} AI in the following tasks"}
\end{itemize}

To ensure that participants understand this information, they were required to spend at least 10 seconds on this information page and confirm their understanding by checking a checkbox.

\subsubsection{Participants}
Participants of EXP2 were recruited online from the Prolific platform, meeting the following criteria: (1) aged between 21 and 60 years, (2) English as their first language, and (3) able to use a personal computer. We ensured gender balance during the recruitment. Each participant could take part in the experiment only once. The experiment was expected to last 25 minutes, with participants compensated at Prolific's recommended rate of £9 per hour. After excluding participants who did not complete the task for various reasons and those who failed multiple attention checks, EXP1 comprised 126 unique participants (42 per condition). Of these, 51.6\% were female, with an average age of 34.0 years ($SD=10.1$), and 73.8\% held at least a bachelor's degree.

\subsection{Results of EXP2}
To address \textbf{RQ3} and \textbf{RQ4}, exploring the effects of communicating AI confidence calibration levels on relevant variables compared to conditions without such communication, this study analyzed differences of relevant variables between participants in EXP2 and EXP1 under the same AI confidence calibration level. This approach followed the comparison method used in previous research \cite{zhang2020effect}.  
To answer {\bf RQ5}, the study examined differences in relevant variables among participants in EXP2 across different AI confidence calibration levels. The specific results are as follows:

\subsubsection{Effect of Communicating AI Confidence Calibration Level on Participants' Detection of AI Confidence Miscalibration (RQ3)}
\begin{figure}[t]
\centering 
\includegraphics[width=0.5\textwidth]{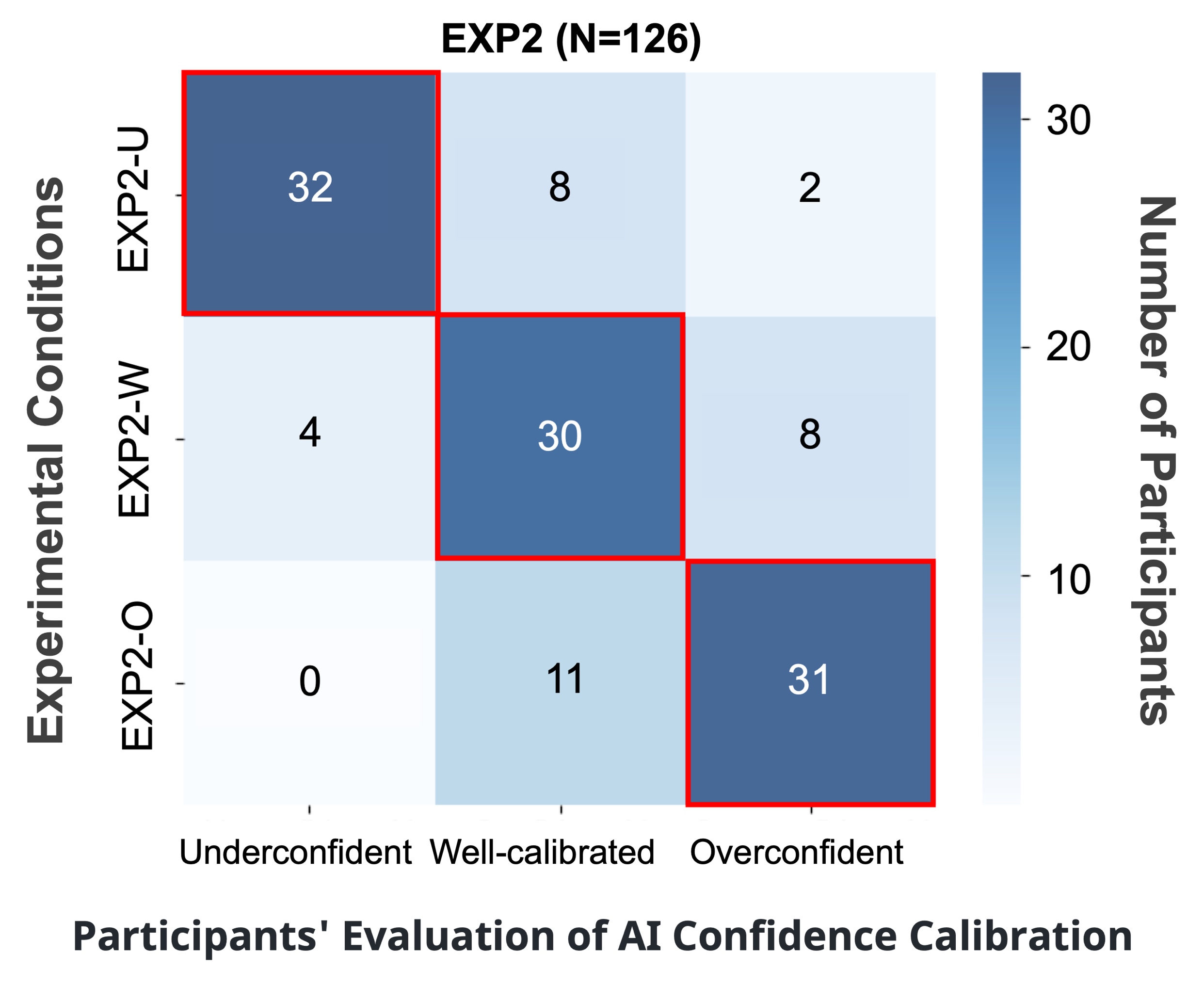} 
\caption{Participants' evaluations of AI confidence calibration level in Experiment 2, shown as a heatmap. The rows represent the three experimental conditions, and the columns indicate participants' evaluations of AI confidence calibration level. The numbers in each cell represent the count of participants giving that particular evaluation in corresponding experimental condition. The darker blue colors indicate a higher number of participants. Correct evaluations of corresponding conditions are highlighted with red boxes.}
\label{heatmap2}
\Description{This figure represents participants' evaluations of AI confidence calibration in Experiment 2, shown as a heatmap. The rows represent the three experimental conditions: the underconfident condition, the well-calirbated condition and the overconfident condition. The columns indicate participants' evaluations of AI confidence calibration: underconfident, well-calibrated, and overconfident. The numbers in each cell represent the count of participants giving that particular evaluation in corresponding experimental condition (from left to right, upper to bottom, the numbers are: 32, 8, 2, 4, 30, 8, 0, 11, 31). The darker blue colors indicate a higher number of participants. Correct evaluations in corresponding conditions are highlighted with red boxes. The underconfident condition underconfident evaluation cell, the well-calibrated condition well-calibrated evaluation cell and the overconfident condition overconfident evaluation cell are highlighted.}
\end{figure}

As shown in Figure \ref{heatmap2}, in EXP2-U, 32 participants (76.2\%) reported that the AI was underconfident, 8 participants (19.0\%) thought the AI was well-calibrated, and 2 participants (4.8\%) considered the AI to be overconfident. The $\chi^2$ test revealed that the distribution of participants' evaluations of AI confidence calibration level in EXP2-U significantly differed from that in EXP1-U ($\chi^2(2, 84) = 20.202$, $p < 0.001$). There were more participants in EXP2-U who correctly identified the AI as underconfident and fewer who mistakenly thought it was well-calibrated. 
In EXP2-O, 31 participants (73.8\%) reported the AI as overconfident, 11 participants (26.2\%) believed it was well-calibrated, and no participants thought it was underconfident. The $\chi^2$ test showed that the distribution of participants' evaluations of AI confidence calibration level in EXP2-O significantly differed from that in EXP1-O ($\chi^2(2, 84) = 20.261$, $p < 0.001$). There were more participants in EXP2-O who correctly recognized the AI as overconfident and fewer who incorrectly thought it was well-calibrated. 
The distribution of participants' evalutations of AI confidence calibration level in EXP2-W did not show a significant difference from that in EXP1-W.
Furthermore, significant differences were observed in the distribution of participants' evaluations of AI confidence calibration leel between EXP2-U and EXP2-W ($\chi^2(2, 84) = 38.115$, $p < 0.001$), as well as between EXP2-O and EXP2-W ($\chi^2(2, 84) = 26.369$, $p < 0.001$).

\textbf{These results indicate that communicating AI confidence calibration levels can help participants better detect AI confidence miscalibration. When participants collaborate with uncalibrated AI, such communication increases the number of participants who correctly identify AI miscalibration. At this stage, for each group in EXP2, most participants were able to correctly estimate the AI's confidence calibration level.}

\subsubsection{Effect of Communicating AI Confidence Calibration Level on Participants' Trust in AI (RQ4.a)}
\label{sec:RQ4.a}
Considering participants in underconfident AI conditions, both EXP1-U and EXP2-U were analyzed. One-way Welch's ANOVA (not passing the Homogeneity test, Levene's test result: $F(1, 82) = 4.571$, $p = 0.035$) showed a significant main effect of communicating AI confidence calibration level on participants' trust in AI confidence score ($F(1, 74.521) = 9.479$, $p = 0.003$, $\eta^2 = 0.104$). Trust in AI confidence score in EXP2-U ($M=4.262$, $SD=1.110$) was significantly lower than in EXP1-U ($M=4.912$, $SD=0.800$). 
Meanwhile, another one-way Welch's ANOVA (not passing the Homogeneity test, Levene's test result: $F(1, 82) = 6.148$, $p = 0.015$) revealed a significant main effect of communicating AI confidence calibration level on participants' trust in AI prediction ($F(1, 73.027) = 9.090$, $p = 0.004$, $\eta^2 = 0.100$). Trust in AI prediction in EXP2-U ($M=4.421$, $SD=0.996$) was significantly lower than in EXP1-U ($M=4.985$, $SD=0.691$).

Considering participants in overconfident AI conditions, both EXP1-O and EXP2-O were analyzed. One-way ANOVA revealed a significant main effect of communicating AI confidence calibration level on participants' trust in AI confidence score ($F(1, 82) = 11.192$, $p = 0.001$, $\eta^2 = 0.120$). Trust in AI confidence score in EXP2-O ($M=4.182$, $SD=1.086$) was significantly lower than in EXP1-O ($M=4.921$, $SD=0.933$). 
At the same time, another one-way ANOVA results showed a significant main effect of communicating AI confidence calibration level on participants' trust in AI prediction ($F(1, 82) = 21.160$, $p < 0.001$, $\eta^2 = 0.205$). Trust in AI prediction in EXP2-O ($M=4.420$, $SD=0.857$) was significantly lower than in EXP1-O ($M=5.215$, $SD=0.721$).

For well-calibrated AI, no significant differences were found in trust in AI prediction ($F(1, 82) = 0.946$, $p=0.334$, $\eta^2=0.011$) or trust in AI confidence score ($F(1, 82) = 0.937$, $p=0.336$, $\eta^2=0.011$) between participants in EXP1-W and EXP2-W.

\begin{figure}[t]
\centering 
\includegraphics[width=0.95\textwidth]{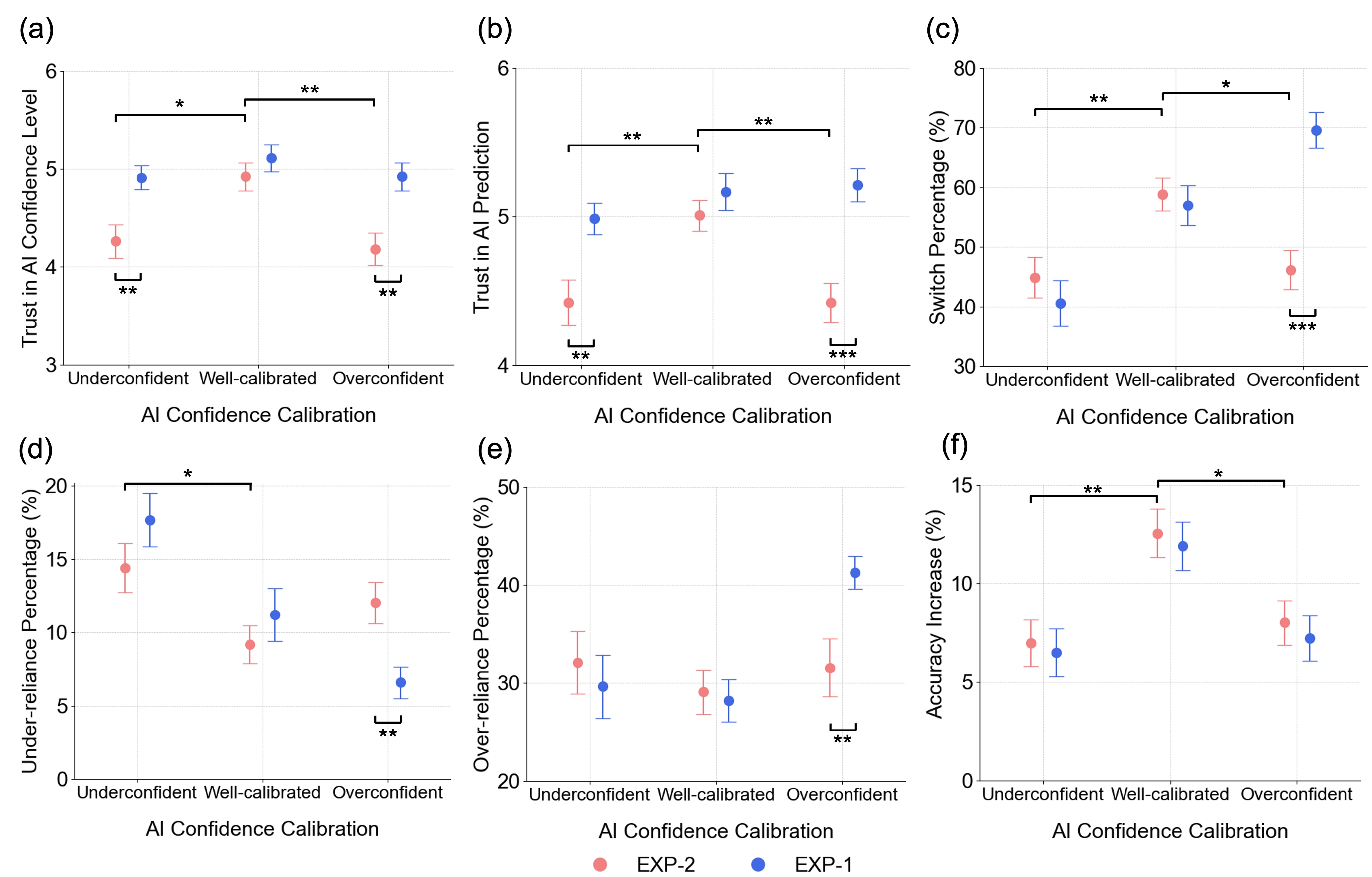} 
\caption{Experiment 2 results across three conditions comparing to experiment 1 results. The dots represent the mean values, and the error bars show one standard error. The significance levels are labeled ($p<0.05$: *, $p<0.01$: **, $p<0.001$: ***). 
{\bf (a)} Trust in AI confidence score.
{\bf (b)} Trust in AI prediction.
{\bf (c)} Switch percentage.
{\bf (d)} Under-reliance percentage.
{\bf (e)} Over-reliance percentage.
{\bf (f)} Accuracy increase.}
\label{exp12_result}
\Description{There are six subfigures in this figure, showing the results of experiment 2 comparing to experiment 1, including trust in AI confidence score, trust in AI prediction, switch percentage, under-reliance percentage, over-reliance percentage, and accuracy increase, as in order. In each subfigure, the dots represent the mean value, and the error bars show one standard error. The significance levels are labeled ($p<0.05$: *, $p<0.01$: **, $p<0.001$: ***). The specific results are given in the results subsection.}
\end{figure}

\textbf{These results indicate that presenting communicating AI confidence calibration level significantly reduces participants' trust in uncalibrated AI, as shown in Fig.~\ref{exp12_result} (a) and (b). This reduction occurs both in their trust in the predictions of uncalibrated AI and their trust in the uncalibrated confidence scores.}

\subsubsection{Effect of Communicating AI Confidence Calibration Level on Participants' Reliance on AI (RQ4.b)}
\label{sec:RQ4.b}
Considering the participants under underconfident AI conditions, i.e., EXP1-U and EXP2-U, one-way ANOVA results showed that the main effects of communicating AI confidence calibration level on switch percentage ($F(1, 82) = 0.733$, $p=0.394$, $\eta^2=0.009$), under-reliance percentage ($F(1, 82) = 1.741$, $p=0.191$, $\eta^2=0.021$), and over-reliance percentage ($F(1, 82) = 0.288$, $p=0.593$, $\eta^2=0.004$) were not significant.

For participants under overconfident AI conditions, i.e., EXP1-O and EXP2-O, one-way ANOVA results indicated a significant main effect of communicating AI confidence calibration level on switch percentage ($F(1, 82) = 28.192$, $p < 0.001$, $\eta^2 = 0.256$). The switch percentage in EXP2-O ($M = 46.137\%$, $SD = 21.185\%$) was significantly lower than in EXP1-O ($M = 69.587\%$, $SD = 19.246\%$). Another one-way ANOVA revealed a significant main effect on under-reliance percentage ($F(1, 82) = 9.515$, $p = 0.003$, $\eta^2 = 0.104$), with a significant higher under-reliance percentage in EXP2-O ($M = 12.018\%$, $SD = 9.062\%$) compared to EXP1-O ($M = 6.576\%$, $SD = 6.973\%$).  Meanwhile, one-way Welch's ANOVA (not passing the Homogeneity test, Levene's test result: $F(1, 82) = 15.190$, $p < 0.001$) showed a significant main effect on over-reliance percentage ($F(1, 64.890) = 8.182$, $p = 0.006$, $\eta^2 = 0.091$), with participants in EXP2-O ($M = 31.560\%$, $SD = 19.112\%$) exhibiting a significantly lower over-reliance percentage than those in EXP1-O ($M = 41.257\%$, $SD = 10.835\%$).

For participants under well-calibrated AI conditions, i.e., EXP1-W and EXP2-W, one-way ANOVA results indicated that the main effects of communicating AI confidence calibration level on switch percentage ($F(1, 82) = 0.178$, $p=0.674$, $\eta^2 =0.002$), under-reliance percentage ($F(1, 82) = 0.845$, $p=0.361$, $\eta^2=0.010$), and over-reliance percentage ($F(1, 82) = 0.074$, $p=0.786$, $\eta^2=0.001$) were not significant.

\textbf{These results suggest that communicating AI confidence calibration level reduces participants' reliance on overconfident AI during collaboration, alleviating over-reliance. However, it increases under-reliance among these participants under overconfident AI conditions, as shown in Fig.~\ref{exp12_result} (c) to (e). Meanwhile, communicating AI confidence calibration level does not mitigate the problem of under-reliance when collaborating with underconfident AI.}

\subsubsection{Effect of Communicating AI Confidence Calibration Level on AI-assisted Decision Making Efficacy (RQ4.c)}
\label{sec:RQ4.c}
No significant main effects of communicating AI confidence calibration level on the accuracy increase after AI advice were observed under any conditions: underconfident AI (EXP1-U and EXP2-U, $F(1, 82) = 0.078$, $p=0.780$, $\eta^2=0.001$), well-calibrated AI (EXP1-W and EXP2-W, $F(1, 82) = 0.133$, $p=0.716$, $\eta^2=0.002$), or overconfident AI (EXP1-O and EXP2-O, $F(1, 82) = 0.243$, $p=0.623$, $\eta^2=0.003$). \textbf{This result suggests that such information can not enhance the efficacy of AI-assisted decision making, as shown in Fig.~\ref{exp12_result} (f).}

\subsubsection{Effect of AI Confidence Miscalibration on Participants' Trust in AI with communicating AI confidence calibration level (RQ5.a)}
\label{sec:RQ5.a}
For participants in EXP2, first, one-way ANOVA revealed that AI confidence calibration had a significant main effect on participants' trust in AI confidence score in EXP2 ($F(2, 123) = 6.388$, $p = 0.002$, $\eta^2 = 0.094$). Post-hoc analysis showed that trust in AI confidence score in EXP2-U ($M=4.262$, $SD=1.110$) was significantly lower ($t = -2.903$, $p = 0.012$) than in EXP2-W ($M = 4.920$, $SD = 0.911$). Trust in AI confidence score in EXP2-O ($M=4.182$, $SD=1.086$) was also significantly lower ($t = -3.258$, $p = 0.004$) than in EXP2-W ($M = 4.920$, $SD = 0.911$). No significant difference was found between EXP2-U and EXP2-O in trust in AI confidence score ($t=0.292$, $p=0.933$).

Meanwhile, one-way Welch's ANOVA (not passing the Homogeneity test, Levene's test result: $F(2, 123) = 3.330$, $p = 0.039$) showed a significant main effect of AI confidence miscalibration on participants' trust in AI predictions ($F(2, 79.975) = 8.122$, $p < 0.001$, $\eta^2 = 0.097$). Post-hoc analysis indicated that trust in AI predictions in EXP2-U ($M=4.421$, $SD=0.996$) was significantly lower ($t = -3.154$, $p = 0.007$) than in EXP2-W ($M = 5.008$, $SD = 0.681$). Trust in AI predictions in EXP2-O ($M=4.420$, $SD=0.857$) was also significantly lower ($t = -3.482$, $p = 0.002$) than in EXP2-W ($M = 5.008$, $SD = 0.681$). No significant difference was observed between EXP2-U and EXP2-O in trust in AI predictions ($t=0.004$, $p=1.000$).

\textbf{These results indicate that in conditions where communicating AI confidence miscalibration level is presented, an uncalibrated AI collaborator reduces participants' trust in them, as shown in Fig.~\ref{exp12_result} (a) and (b). This reduction is not limited to trust in AI confidence scores but also affects participants' trust in AI predictions.}

\subsubsection{Effect of AI Confidence Miscalibration on Participants' Reliance on AI with Communicating AI Confidence Calibration Level (RQ5.b)}
\label{sec:RQ5.b}
Considering participants' reliance on AI in EXP2, first focusing on their switch percentage, one-way ANOVA results indicated a significant main effect of AI confidence miscalibration on participants' switch percentage in EXP2 ($F(2, 123) = 5.997$, $p = 0.003$, $\eta^2 = 0.089$). Post-hoc analysis showed that the switch percentage in EXP2-U ($M=44.859\%$, $SD=21.896\%$) was significantly lower ($t=-3.132$, $p=0.006$) than in EXP2-W ($M=58.818\%$, $SD=17.967\%$). The switch percentage in EXP2-O ($M = 46.137\%$, $SD = 21.185\%$) was also significantly lower ($t=-2.846$, $p=0.014$) than in EXP2-W ($M=58.818\%$, $SD=17.967\%$). No significant difference was found between EXP2-U and EXP2-O in switch percentage ($t=-0.287$, $p=0.956$).

For the under-reliance percentage in EXP2, one-way ANOVA results showed a significant main effect of AI confidence calibration ($F(2, 123) = 3.168$, $p = 0.046$, $\eta^2 = 0.049$). Post-hoc analysis indicated that the under-reliance percentage in EXP2-U ($M=14.399\%$, $SD=10.904\%$) was significantly higher ($t=2.514$, $p=0.035$) than in EXP2-W ($M=9.184\%$, $SD=8.374\%$). Although the under-reliance percentage in EXP2-O ($M = 12.018\%$, $SD = 9.062\%$) was numerically higher than in EXP2-W ($M=9.184\%$, $SD=8.374\%$), this difference was not statistically significant ($t=1.366$, $p=0.362$). No significant difference was observed between EXP2-U and EXP2-O in under-reliance percentage ($t=1.148$, $p=0.487$). 
Additionally, linear regression results showed a significant negative linear correlation between participants' under-reliance percentage and switch percentage ($r=-0.680$, $p<0.001$).

For the over-reliance percentage in EXP2, one-way ANOVA results showed that the main effect of AI confidence miscalibration was not significant ($F(2,123)=0.323$, $p=0.725$, $\eta^2=0.005$).

\textbf{These results suggest that in conditions where communicating AI confidence calibration level is presented, uncalibrated AI reduces human reliance on it and leads to higher participants' under-reliance, as shown in Fig.~\ref{exp12_result} (c) to (e). }

\subsubsection{Effect of AI Confidence Miscalibration on AI-assisted Decision Making Efficacy with Communicating AI Confidence Calibration Level (RQ5.c)}
\label{sec:RQ5.c}
For the increase of accuracy after AI advice in EXP2, as shown in Fig.~\ref{exp12_result} (f), one-way ANOVA results indicated a significant main effect of AI confidence miscalibration ($F(2, 123) = 6.217$, $p = 0.003$, $\eta^2 = 0.092$). Post-hoc analysis revealed that the increase in accuracy in EXP2-U ($M = 6.984\%$, $SD = 7.715\%$) was significantly lower ($t = -3.315$, $p = 0.003$) than in EXP2-W ($M = 12.540\%$, $SD = 7.990\%$). The increase in accuracy in EXP2-O ($M = 8.015\%$, $SD = 7.326\%$) was also significantly lower ($t = -2.669$, $p = 0.021$) than in EXP2-W ($M = 12.540\%$, $SD = 7.990\%$). No significant difference was found between the increases in accuracy in EXP2-U and EXP2-O ($t=-0.615$, $p=0.812$).

Linear regression results indicated that in EXP2, participants' under-reliance percentage had a significant negative linear correlation with the increase in accuracy ($r = -0.271$, $p = 0.002$). Similarly, participants' over-reliance percentage also showed a significant negative linear correlation with the increase in accuracy ($r = -0.494$, $p < 0.001$).

\textbf{These results suggest that in conditions where communicating AI confidence calibration level is presented, uncalibrated AI still weakens the efficacy of AI-assisted decision making, which is associated with participants' inappropriate reliance (both over-reliance and under-reliance).}

\section{Discussion}

This study explores users' awareness of AI confidence miscalibration and reveals the effect of AI confidence miscalibration on humans' trust and reliance on AI, as well as the efficacy of collaboration. 
Our findings indicate that most participants are unable to correctly detect AI miscalibration. Then they tend to overly rely on overconfident AI and insufficiently rely on underconfident AI, ultimately reducing the efficacy of AI-assisted decision making. 
We also discovered that communicating AI confidence calibration level could help people better detect AI miscalibration and mitigate the tendency to overly rely on overconfident AI. However, it reduced users' trust in uncalibrated AI, manifesting as users' under-reliance on them. Merely providing the information did not aid in improving the efficacy of AI-assisted decision making.
In this section, we summarize the aforementioned findings and discuss the underlying reasons for these phenomena. Next, we outline the theoretical and practical implications of our study. Finally, we explore potential directions for future work.

\subsection{The Challenges Faced by Users in Detecting AI Confidence Miscalibration}
We found that users were facing challenges in detecting AI confidence miscalibration. These challenges can be attributed primarily to two factors. First, identifying uncalibrated AI is inherently complex. Unlike past studies where users could evaluate AI accuracy by observation \cite{rechkemmer2022confidence}, recognizing AI confidence calibration level during interaction requires accurate observation of both AI accuracy and confidence scores, as well as the ability to assess their correspondence. For general users without a background in AI and machine learning, this is undoubtedly challenging.
Secondly, another possible reason is that users are generally unfamiliar with the concept of AI confidence calibration. While users may grasp the literal term "confidence calibration," they are unclear about its precise definition. Like many other AI-related terms \cite{lai2021towards}, it is not commonly encountered in everyday life. Although we introduced the definitions of calibration-related terminology to the participants in this study, we cannot expect those encountering these definitions for the first time to fully understand and apply them accurately.

Interestingly, many participants regarded the AI as well-calibrated despite interacting with overconfident or underconfident AI models. We believe this reflects a strategic approach by participants in evaluating AI capabilities; they default to assuming functionality is sound when specific shortcomings are not explicitly discernible. 
Meanwhile, a minority of participants were able to correctly identify miscalibrations, while a few participants provided evaluations that were completely contrary to the actual AI confidence calibration, highlighting differences in participants' cognitive abilities and evaluation capacities regarding AI confidence calibration.

\subsection{Communicating AI Confidence Calibration Level is a Double Edged Sword}
We find that communicating AI confidence calibration level can help many participants recognize AI miscalibration. This suggests that most participants adopted our statements about AI performance after observing AI confidence scores during the Training Phase, consistent with findings from past research on providing information about AI models \cite{lai2021towards,rechkemmer2022confidence}. This adoption may be due to the alignment between participants' observations and our statements, or it could stem from an underlying anchoring effect \cite{furnham2011literature}. 
Some participants did not adopt our statements but stuck to their own judgments, reflecting skepticism among some participants regarding statements about AI performance, or possibly due to overconfidence in their own judgments.

However, communicating AI confidence calibration also has undeniable drawbacks. When participants became aware that the AI was uncalibrated, they lost trust in its confidence scores. 
We believe the diminished trust represents an attitudinal shift upon realizing a flaw in the AI, stemming also from a disappointment due to the AI's performance deviating from participants' default expectations (that the AI is well-calibrated). Additionally, some participants might perceive the AI's overconfidence or underconfidence as a form of dishonesty or deliberate deception, potentially leading to further erosion of trust in the AI. However, due to the lack of further qualitative data as evidence, this remains a speculative hypothesis. 
Interestingly, at the same tine, participants' trust in the AI's ability to make correct predictions also diminished.
The cascading decline in trust in AI predictions reflects a potential reality that participants' trust in different aspects of AI's capabilities is interconnected and not independent. This also support viewpoints from past research that users' trust is a complex concept and a multidimensional construct, with potential intrinsic connections between different trust measurements \cite{rechkemmer2022confidence, lee2004trust}.

The decline in trust subsequently influences participants' reliance on AI. For participants collaborating with underconfident AI, we expected that communicating AI confidence calibration level would help participants adjust their acceptance strategy of AI suggestions, thus enhancing their reliance on AI and reducing levels of under-reliance. Contrary to our expectations, due to the decrease in trust, these participants did not show a significant increase in reliance on AI, nor was there a noticeable improvement in under-reliance. For groups working with overconfident AI, we expected that communicating AI confidence calibration level would reduce users' reliance and over-reliance levels. In this respect, we were successful. However, an unforeseen consequence was that due to the decrease in trust, participants' reliance on AI over-decreased, resulting in increased levels of under-reliance. These phenomena reflect the effects of human trust attitudes on reliance behaviors \cite{lee2004trust}. Since participants exhibited relative high levels of under-reliance in both conditions, that is, a greater rejection of correct AI suggestions, communicating AI confidence calibration level did not enhance the efficacy of uncalibrated AI collaboration with humans as expected.

\subsection{AI Confidence Miscalibration Hinders AI-assisted Decision Making}
Past research suggests that as long as AI confidence is well-calibrated, it can help people rely on AI appropriately \cite{lai2021towards}; we further supplement this statement by adding that if AI confidence is uncalibrated, it messes users' appropriate reliance up and invariably hinders AI-assisted decision making.
Without communicating AI confidence calibration level, participants relied more on overconfident AI and tended to over-rely on it; or they relied less on underconfident AI and exhibited under-reliance on it. This is because when AI confidence scores are higher, people are more inclined to adopt AI suggestions, and vice versa \cite{zhang2020effect}. These inappropriate reliance behaviors, in turn, lead to a decline in efficacy.
This outcome aligns with analyses from a previous study, which highlighted issues arising from inconsistencies between AI confidence and correctness \cite{ma2024you}. 
Meanwhile, AI confidence miscalibration did not significantly impact participants' trust in AI without communicating AI confidence calibration level. We think it is because participants did not recognize the AI's miscalibration. This aligns with one previous study's results, which indicate that AI's confidence score does not significantly affect human post-collaboration reported trust in AI \cite{rechkemmer2022confidence}, but it is opposite to another previous study which suggests contrary results\cite{yin2019understanding}.
When information about AI confidence calibration was provided, participants' distrust towards uncalibrated AI leads them to under-rely on it, resulting in lower AI-assisted decision making efficacy compared to when working with well-calibrated AI.
These findings highlight the hazards of uncalibrated AI and how it causes these hazards, also emphasizing that calibrating AI confidence is one of the vital tasks at present.

\subsection{Theoretical and Design Implications}
This study has significant theoretical implications as it fills a research gap in the existing literature on AI-assisted decision making regarding the effects of AI confidence miscalibration and the insufficient discussion about users' understanding of it \cite{lai2021towards,zhang2020effect,ma2024you}. It offers insights into how AI confidence miscalibration affects human decision-makers' trust, reliance, and the efficacy of AI-assisted decision making. This study highlights the risks associated with uncalibrated AI in AI-assisted decision making.
Furthermore, this research treats AI confidence calibration level as a type of performance information provided to users, enriching studies on improving AI-assisted decision making through the provision of model information \cite{rechkemmer2022confidence,lai2021towards,lai2019human}.
It reveals how such information affects users' attitudes and behaviors.

For design implications, the inadequate detection of AI confidence miscalibration among users reveals a practical design need: when aiming to promote users' understanding of AI's error boundaries by displaying AI confidence scores, interventions must also be taken to help users comprehend AI confidence. More specifically, users need to know not only what AI confidence scores represent but also whether they are reliable (calibrated). However, stopping there is insufficient, as our findings indicate that if users are aware that AI confidence scores are unreliable, they will lose trust in AI. Facing situations with uncalibrated AI, an ideal optimization would be for users to understand the distribution characteristics of uncalibrated AI's confidence scores, infer the true probability of correctness based on the AI-provided confidence score, and appropriately rely on AI. Therefore, AI developers need to further inform users about why and under what characteristics AI confidence scores might be unreliable, and what should be done about it. 
How to present such information about models \cite{rechkemmer2022confidence,lai2021towards} to users in a simple and understandable way will test the wisdom of future developers.
Another “shortcut” to address this issue would be to simply not display its confidence score to users when AI is uncalibrated. We do not recommend this approach as it essentially avoids addressing the problem.

\subsection{Uncalibrated AI Confidence: Defects, Dishonesty or Deception}
From the current findings, a broader question arises regarding how we should perceive the nature of uncalibrated AI confidence, which yields different results from different perspectives. Purely from a technical standpoint, miscalibration is a technical challenge or flaw. However, from the perspective of honesty, the overconfidence or underconfidence of AI could be seen as a form of dishonest behavior, exaggerating or downplaying its normal confidence score, potentially deceiving users and manipulating their behavior. Of course, this interpretation pertains only to the behavior itself, not to motives; in our study, the AI was not programmed with the intent to lie or manipulate.

Humans can use their own overconfidence to deceive \cite{schwardmann2019deception}, and worryingly, our study along with some past research suggests that AI might also deceive and manipulate humans by adjusting its confidence scores \cite{zhang2020effect,schneider2020deceptive,park2024ai}; similarly, third parties manipulating users through AI confidence is also possible. For example, in a poker game, an AI might bluff to intimidate other players into folding \cite{brown2019superhuman}. Such manipulations relying on confidence, if used maliciously with technology, are extremely dangerous, being more covert and difficult to detect than direct commands and recommendations. For instance, in critical decision-making scenarios such as investments, manipulation through confidence could sway users towards or away from certain decisions, benefiting from the users' resultant losses. Even in elections, by exaggerating the confidence in certain information, AI might manipulate and distort people’s electoral preferences, posing greater risks.

As AI continues to evolve and permeate human life, this risk becomes more severe. Facing these serious dangers, both practitioners and policymakers should assume their respective responsibilities \cite{park2024ai, schneider2020deceptive}. Before deploying AI models, developers should disclose information related to model calibration in a timely manner; policymakers should also promote the establishment of related laws and industry standards, and provide more policy support and funding for research related to calibrating AI and monitoring AI risks. Moreover, users should develop an understanding of these risks and remain vigilant against the manipulation of their behavior by AI or third parties using AI confidence. In summary, a multifaceted effort is required to address this risk, and this paper calls for future research to engage in a broader and more in-depth discussion of this issue.

\subsection{Limitations and Future Work}
There are several limitations of our work. 
In this experiment, we utilized the confidence score as a form of expressing AI's uncertainty. Beyond the most basic numerical expression of AI uncertainty, there are other forms of representing AI uncertainty, such as categorizing AI confidence from high to low in discrete classes, and the verbalization methods commonly used in large language models research \cite{xiong2023llms}. It would be valuable to explore whether the impact of AI uncalibrated confidence and individuals' perception of AI confidence quality differ under these various representations compared to the confidence score form. This is particularly important today, as more advanced AI models like large language models are being widely used by users.
At the same time, compared to the overconfident and underconfident conditions designed in this experiment, the miscalibration issues AI faces in practice are more complex and diversified. For example, the same AI model may be overconfident about some issues and underconfident about others, exhibiting mixed miscalibration characteristics \cite{guo2017calibration}. In the context of users struggling to perceive miscalibration, the variability of these miscalibration patterns could pose greater risks and challenges to AI-assisted decision making.

Furthermore, the generalizability of this study is limited to the category of generic AI-assisted decision making scenarios represented by city image recognition.
In high-risk scenarios such as in healthcare and military applications, people's decision making may lean towards conservatism based on Loss Aversion and Prospect Theory \cite{tversky1991loss, kahneman2013prospect}. In such contexts, individuals may exhibit behaviors different from those observed in our experiments. AI uncalibrated confidence might pose greater risks, and its specific impacts warrant further investigation.
Lastly, since our participants were general users recruited online, caution should be exercised when generalizing the findings to other groups. Consideration should be given to the differences in AI literacy among different groups. For example, AI industry professionals might perform better in collaborations with uncalibrated AI compared to general users, while groups like the elderly, technophobes, and students may be influenced by their varying familiarity with AI \cite{ng2021conceptualizing, ng2022using,kasinidou2023promoting}.

Building on our results, a important future task is to further understand how users perceive and understand AI confidence calibration level. It is crucial to explore additional methods to enhance users' understanding and perception of AI confidence calibration level, thereby improving their AI literacy \cite{ng2021conceptualizing}, which is necessary for more effective human-AI collaboration. Meanwhile, further exploration of the risks associated with uncalibrated AI under different forms of AI uncertainty representation, and whether the influences differ across various scenarios and demographic groups, would help researchers comprehensively understand the risks of miscalibration and establish a robust theoretical framework. Moreover, investigating how users develop trust in different aspects of AI capabilities and their interrelationships is a meaningful future endeavor that can guide the development of more trustworthy AI and foster more harmonious human-AI relations. Lastly, we also call for more research focused on how to disclose, monitor, and ultimately resolve issues of AI model miscalibration, to control and eliminate the risks and challenges that miscalibration brings.

\section{Conclusions}
Faced with the widespread issue of AI confidence miscalibration, it becomes crucial to explore the influence of AI confidence miscalibration and users' awareness of AI confidence miscalibration. In this work, through two experiments, we empirically showed: 1) users faced challenges in detecting AI confidence miscalibration; 2) AI confidence miscalibration could lead to either over-reliance or under-reliance on AI systems; 3) while communicating AI confidence calibration levels could improve users' detection of AI miscalibration, it simultaneously eroded users' trust in uncalibrated AI, resulting in high levels of under-reliance on uncalibrated AI and not improved decision efficacy.
These findings highlight the significance and challenges of AI confidence calibration in AI-assisted decision-making. They also suggest that merely communicating calibration level with users is insufficient to enhance decision-making efficacy. 
Future efforts should focus on developing effective strategies for communicating AI confidence scores to foster users' appropriate trust and reliance on AI, aiming to achieve optimal outcomes in AI-assisted decision-making.
We hope our research will inspire more discussion in this direction and raise awareness of the ethical risks potentially associated with AI miscalibration.


\bibliographystyle{ACM-Reference-Format}
\bibliography{sample-base}

\appendix
\newpage
\section{Survey Items}
\label{app: survey}
\subsection{Evaluation of AI Confidence Calibration}
Participants were asked to report their evaluation of the AI confidence calibration, choosing between {\it Underconfident}, {\it Well-calibrated}, or {\it Overconfident}. When answering questions, participants first received descriptions of three AI confidence calibration conditions:
\begin{itemize}
    \item When the AI's average confidence level is lower than the accuracy of its predictions, we term it 'underconfident.' 
    \item When the AI's average confidence level aligns closely with the accuracy of its predictions, we describe it as 'well-calibrated.' 
    \item When the AI's average confidence level exceeds the accuracy of its predictions, we label this 'overconfident.' 
\end{itemize}
Then, they were asked to assess the AI confidence calibration:
\begin{itemize}
    \item \textit{Please evaluate the AI confidence calibration among previous tasks.}
\end{itemize}

\subsection{Trust in AI Prediction}
Participants' trust in AI prediction was measured by a 7-point Likert scale  (1="Strongly disagree",7="Strongly agree"; $M=4.869$, $SD=0.856$; McDonald's $\omega=0.796$, 95\% CI: [0.754, 0.839]). Here are the questions:
\begin{enumerate}
    \item \textit{I trust the AI's predictions.} 
    \item \textit{I am comfortable adopting the AI's predictions.} 
    \item \textit{I believe the AI uses appropriate methods to reach predictions.} 
\end{enumerate}

\subsection{Trust in AI Confidence Level}
Participants' trust in AI confidence level was measured by a 7-point Likert scale (1="Strongly disagree",7="Strongly agree"; $M=4.718$, $SD=1.018$; McDonald's $\omega=0.862$, 95\% CI: [0.833, 0.892]). Here are the questions:
\begin{enumerate}
    \item \textit{I trust the AI's confidence levels in its predictions.} 
    \item \textit{I am comfortable referring to the AI's confidence levels.} 
    \item \textit{I believe the AI uses reasonable methods to evaluate confidence levels.} 
\end{enumerate}
\newpage
\section{Statistical Analysis Results}

\begin{table}[h]
\caption{Descriptive statistics of conditions in Experiment 1.}
\begin{tabular}{cccc}
\hline
\textbf{Dependent Variable} & \textbf{Experimental Group} & \textbf{Mean} & \textbf{S.D.} \\ \hline
\multirow{3}{*}{Trust in AI Confidence Level} & EXP1-U & 4.912    & 0.800    \\
                                              & EXP1-W & 5.111    & 0.895    \\
                                              & EXP1-O & 4.921    & 0.933    \\ \hline
\multirow{3}{*}{Trust in AI Prediction}       & EXP1-U & 4.985    & 0.691    \\
                                              & EXP1-W & 5.166    & 0.804    \\
                                              & EXP1-O & 5.215    & 0.721    \\ \hline
\multirow{3}{*}{Switch Percentage}            & EXP1-U & 40.492\% & 24.752\% \\
                                              & EXP1-W & 56.975\% & 21.868\% \\
                                              & EXP1-O & 69.587\% & 19.246\% \\ \hline
\multirow{3}{*}{Under-reliance Percentage}    & EXP1-U & 17.687\% & 11.912\% \\
                                              & EXP1-W & 11.224\% & 11.703\% \\
                                              & EXP1-O & 6.576\%  & 6.973\%  \\ \hline
\multirow{3}{*}{Over-reliance Percentage}     & EXP1-U & 29.632\% & 20.953\% \\
                                              & EXP1-W & 28.207\% & 14.002\% \\
                                              & EXP1-O & 41.257\% & 10.835\% \\ \hline
\multirow{3}{*}{Accuracy Increase}            & EXP1-U & 6.508\%  & 7.859\%  \\
                                              & EXP1-W & 11.905\% & 7.967\%  \\
                                              & EXP1-O & 7.211\%  & 7.433\%  \\ \hline
\end{tabular}
\end{table}

\begin{table}[h]
\caption{Descriptive statistics of conditions in Experiment 2.}
\begin{tabular}{cccc}
\hline
\textbf{Dependent Variable} & \textbf{Experimental Group} & \textbf{Mean} & \textbf{S.D.} \\ \hline
\multirow{3}{*}{Trust in AI Confidence Level} & EXP2-U & 4.262    & 1.110    \\
                                              & EXP2-W & 4.920    & 0.911    \\
                                              & EXP2-O & 4.182    & 1.086    \\ \hline
\multirow{3}{*}{Trust in AI Prediction}       & EXP2-U & 4.421    & 0.996    \\
                                              & EXP2-W & 5.008    & 0.681    \\
                                              & EXP2-O & 4.420    & 0.857    \\ \hline
\multirow{3}{*}{Switch Percentage}            & EXP2-U & 44.859\% & 21.896\% \\
                                              & EXP2-W & 58.818\% & 17.967\% \\
                                              & EXP2-O & 46.137\% & 21.185\% \\ \hline
\multirow{3}{*}{Under-reliance Percentage}    & EXP2-U & 14.399\% & 10.904\% \\
                                              & EXP2-W & 9.184\%  & 8.374\%  \\
                                              & EXP2-O & 12.018\% & 9.062\%  \\ \hline
\multirow{3}{*}{Over-reliance Percentage}     & EXP2-U & 32.067\% & 20.598\% \\
                                              & EXP2-W & 29.064\% & 14.806\% \\
                                              & EXP2-O & 31.560\% & 19.112\% \\ \hline
\multirow{3}{*}{Accuracy Increase}            & EXP2-U & 6.984\%  & 7.715\%  \\
                                              & EXP2-W & 12.540\% & 7.990\%  \\
                                              & EXP2-O & 8.015\%  & 7.326\%  \\ \hline
\end{tabular}
\end{table}

\begin{table}[h]
\caption{Assumption checks of the ANOVA about the main effects of AI confidence calibration in Experiment 1. The significance levels are labeled ($p<0.05$: *, $p<0.01$: **, $p<0.001$: ***).}
\begin{tabular}{ccccc}
\hline
\multirow{2}{*}{\textbf{Dependent Variable}} & \multicolumn{4}{c}{\textbf{Levene's Test Results}}                              \\ \cline{2-5} 
                                             & \textbf{F} & \textbf{df (Group)} & \textbf{df (Residual)} & \textit{\textbf{p}}        \\ \hline
Trust in AI Confidence Level & 1.081 & 2 & 123 & 0.342   \\
Trust in AI Prediction       & 0.793 & 2 & 123 & 0.455   \\
Switch Percentage            & 2.555 & 2 & 123 & 0.082   \\
Under-reliance Percentage    & 5.251 & 2 & 123 & 0.006** \\
Over-reliance Percentage                     & 7.892      & 2                   & 123                    & \textless{}0.001*** \\
Accuracy Increase            & 0.049 & 2 & 123 & 0.952   \\ \hline
\end{tabular}
\end{table}

\begin{table}[h]
\caption{Assumption checks of the ANOVA about the main effects of AI confidence calibration in Experiment 2. The significance levels are labeled ($p<0.05$: *, $p<0.01$: **, $p<0.001$: ***).}
\begin{tabular}{ccccc}
\hline
\multirow{2}{*}{\textbf{Dependent Variable}} & \multicolumn{4}{c}{\textbf{Levene's Test Results}}                       \\ \cline{2-5} 
                                             & \textbf{F} & \textbf{df (Group)} & \textbf{df (Residual)} & \textit{\textbf{p}} \\ \hline
Trust in AI Confidence Level & 2.099 & 2 & 123 & 0.127  \\
Trust in AI Prediction       & 3.330  & 2 & 123 & 0.039* \\
Switch Percentage            & 0.597 & 2 & 123 & 0.552  \\
Under-reliance Percentage    & 1.371 & 2 & 123 & 0.258  \\
Over-reliance Percentage     & 1.228 & 2 & 123 & 0.296  \\
Accuracy Increase            & 0.044 & 2 & 123 & 0.957  \\ \hline
\end{tabular}
\end{table}

\begin{table}[h]
\caption{One-way ANOVA results about the main effects of AI confidence calibration in Experiment 1. The significance levels are labeled ($p<0.05$: *, $p<0.01$: **, $p<0.001$: ***).}
\resizebox{\textwidth}{!}{%
\begin{tabular}{ccccccc}
\hline
\multirow{2}{*}{\textbf{Dependent Variable}} & \multirow{2}{*}{\textbf{Statistical Test}} & \multicolumn{5}{c}{\textbf{Main Effect of AI Confidence Calibration}} \\ \cline{3-7} 
                             &               & \textbf{F} & \textbf{df (Group)} & \textbf{df (Residual)} & \textit{\textbf{p}}        & \textbf{$\eta^2$} \\ \hline
Trust in AI Confidence Level & ANOVA         & 0.690       & 2                   & 123                    & 0.503               & 0.011             \\
Trust in AI Prediction       & ANOVA         & 1.127      & 2                   & 123                    & 0.327               & 0.018             \\
Switch Percentage            & ANOVA         & 18.354     & 2                   & 123                    & \textless{}0.001*** & 0.230             \\
Under-reliance Percentage    & Welch's ANOVA & 13.886     & 2                   & 76.379                 & \textless{}0.001*** & 0.163             \\
Over-reliance Percentage     & Welch's ANOVA & 13.156     & 2                   & 77.508                 & \textless{}0.001*** & 0.122             \\
Accuracy Increase            & ANOVA         & 6.001      & 2                   & 123                    & 0.003**             & 0.089             \\ \hline
\end{tabular}
}
\end{table}

\begin{table}[h]
\caption{One-way ANOVA results about the main effects of AI confidence calibration in Experiment 2. The significance levels are labeled ($p<0.05$: *, $p<0.01$: **, $p<0.001$: ***).}
\resizebox{\textwidth}{!}{%
\begin{tabular}{ccccccc}
\hline
\multirow{2}{*}{\textbf{Dependent Variable}} & \multirow{2}{*}{\textbf{Statistical Test}} & \multicolumn{5}{c}{\textbf{Main Effect of AI Confidence Calibration}} \\ \cline{3-7} 
                             &               & \textbf{F} & \textbf{df (Group)} & \textbf{df (Residual)} & \textit{\textbf{p}}        & \textbf{$\eta^2$} \\ \hline
Trust in AI Confidence Level & ANOVA         & 6.388      & 2                   & 123                    & 0.002**             & 0.094             \\
Trust in AI Prediction       & Welch's ANOVA & 8.122      & 2                   & 79.975                 & \textless{}0.001*** & 0.097             \\
Switch Percentage            & ANOVA         & 5.997      & 2                   & 123                    & 0.003**             & 0.089             \\
Under-reliance Percentage    & ANOVA         & 3.168      & 2                   & 123                    & 0.046*              & 0.049             \\
Over-reliance Percentage     & ANOVA         & 0.323      & 2                   & 123                    & 0.725               & 0.005             \\
Accuracy Increase            & ANOVA         & 6.217      & 2                   & 123                    & 0.003**             & 0.092             \\ \hline
\end{tabular}
}
\end{table}

\begin{table}[h]
\caption{Post-hoc analysis results about the effects of AI confidence calibration in Experiment 1. The significance levels are labeled ($p<0.05$: *, $p<0.01$: **, $p<0.001$: ***).}
\resizebox{\textwidth}{!}{%
\begin{tabular}{clccccc}
\hline
\multirow{2}{*}{\textbf{Dependent Variable}} & \multicolumn{6}{c}{\textbf{Post-hoc Analysis Result}}                                   \\ \cline{2-7} 
 &
  \multicolumn{1}{c}{\textbf{Statistical Test}} &
  \multicolumn{2}{c}{\textbf{Experimental Group}} &
  \textbf{t} &
  \textit{\textbf{p}} &
  \textbf{Cohen's d} \\ \hline
\multirow{3}{*}{Trust in AI Confidence Level} &
  \multirow{3}{*}{Tukey HSD} &
  EXP1-U &
  EXP1-W &
  -1.039 &
  0.554 &
  -0.227 \\
                                             &                               & EXP1-O & EXP1-W & -0.994 & 0.582               & -0.217 \\
                                             &                               & EXP1-U & EXP1-O & -0.045 & 0.999               & -0.010 \\ \hline
\multirow{3}{*}{Trust in AI Prediction}      & \multirow{3}{*}{Tukey HSD}    & EXP1-U & EXP1-W & -1.124 & 0.501               & -0.245 \\
                                             &                               & EXP1-O & EXP1-W & 0.301  & 0.951               & 0.066  \\
                                             &                               & EXP1-U & EXP1-O & -1.424 & 0.332               & -0.311 \\ \hline
\multirow{3}{*}{Switch Percentage}           & \multirow{3}{*}{Tukey HSD}    & EXP1-U & EXP1-W & -3.422 & 0.002**             & -0.747 \\
                                             &                               & EXP1-O & EXP1-W & 2.619  & 0.027*              & 0.571  \\
                                             &                               & EXP1-U & EXP1-O & -6.041 & \textless .001***   & 1.318  \\ \hline
\multirow{3}{*}{Under-reliance Percentage}   & \multirow{3}{*}{Games-Howell} & EXP1-U & EXP1-W & 2.508  & 0.037*              & -      \\
                                             &                               & EXP1-O & EXP1-W & -2.211 & 0.077               & -      \\
                                             &                               & EXP1-U & EXP1-O & 5.217  & \textless{}0.001*** & -      \\ \hline
\multirow{3}{*}{Over-reliance Percentage}    & \multirow{3}{*}{Games-Howell} & EXP1-U & EXP1-W & 0.367  & 0.929               & -      \\
                                             &                               & EXP1-O & EXP1-W & 4.777  & \textless .001***   & -      \\
                                             &                               & EXP1-U & EXP1-O & -3.194 & 0.006**             & -      \\ \hline
\multirow{3}{*}{Accuracy Increase}           & \multirow{3}{*}{Tukey HSD}    & EXP1-U & EXP1-W & -3.189 & 0.005**             & -0.696 \\
                                             &                               & EXP1-O & EXP1-W & -2.767 & 0.018*              & -0.604 \\
                                             &                               & EXP1-U & EXP1-O & -0.421 & 0.907               & -0.092 \\ \hline
\end{tabular}
}
\end{table}

\begin{table}[h]
\caption{Post-hoc analysis results about the effects of AI confidence calibration in Experiment 2. The significance levels are labeled ($p<0.05$: *, $p<0.01$: **, $p<0.001$: ***).}
\resizebox{\textwidth}{!}{%
\begin{tabular}{clccccc}
\hline
\multirow{2}{*}{\textbf{Dependent Variable}} & \multicolumn{6}{c}{\textbf{Post-hoc Analysis Result}}                       \\ \cline{2-7} 
 &
  \multicolumn{1}{c}{\textbf{Statistical Test}} &
  \multicolumn{2}{c}{\textbf{Experimental Group}} &
  \textbf{t} &
  \textit{\textbf{p}} &
  \textbf{Cohen's d} \\ \hline
\multirow{3}{*}{Trust in AI Confidence Level} &
  \multirow{3}{*}{Tukey HSD} &
  EXP2-U &
  EXP2-W &
  -2.903 &
  0.012* &
  -0.633 \\
                                             &                               & EXP2-O & EXP2-W & -3.258 & 0.004** & -0.711 \\
                                             &                               & EXP2-U & EXP2-O & 0.292  & 0.933   & 0.077  \\ \hline
\multirow{3}{*}{Trust in AI Prediction}      & \multirow{3}{*}{Games-Howell} & EXP2-U & EXP2-W & -3.154 & 0.007** & -      \\
                                             &                               & EXP2-O & EXP2-W & -3.482 & 0.002** & -      \\
                                             &                               & EXP2-U & EXP2-O & 0.004  & 1.000   & -      \\ \hline
\multirow{3}{*}{Switch Percentage}           & \multirow{3}{*}{Tukey HSD}    & EXP2-U & EXP2-W & -3.132 & 0.006** & -0.684 \\
                                             &                               & EXP2-O & EXP2-W & -2.846 & 0.014*  & -0.621 \\
                                             &                               & EXP2-U & EXP2-O & -0.287 & 0.956   & -0.063 \\ \hline
\multirow{3}{*}{Under-reliance Percentage}   & \multirow{3}{*}{Tukey HSD}    & EXP2-U & EXP2-W & 2.514  & 0.035*  & 0.549  \\
                                             &                               & EXP2-O & EXP2-W & 1.366  & 0.362   & 0.298  \\
                                             &                               & EXP2-U & EXP2-O & 1.148  & 0.487   & 0.250  \\ \hline
\multirow{3}{*}{Over-reliance Percentage}    & \multirow{3}{*}{Tukey HSD}    & EXP2-U & EXP2-W & 0.750  & 0.734   & 0.164  \\
                                             &                               & EXP2-O & EXP2-W & 0.624  & 0.807   & 0.136  \\
                                             &                               & EXP2-U & EXP2-O & 0.127  & 0.991   & 0.028  \\ \hline
\multirow{3}{*}{Accuracy Increase}           & \multirow{3}{*}{Tukey HSD}    & EXP2-U & EXP2-W & -3.315 & 0.003** & -0.723 \\
                                             &                               & EXP2-O & EXP2-W & -2.669 & 0.021*  & -0.589 \\
                                             &                               & EXP2-U & EXP2-O & -0.615 & 0.812   & -0.134 \\ \hline
\end{tabular}
}
\end{table}

\begin{table}[h]
\caption{Assumption checks of the ANOVA about the main effects of information about AI confidence calibration in underconfident conditions. The significance levels are labeled ($p<0.05$: *, $p<0.01$: **, $p<0.001$: ***).}
\resizebox{.7\textwidth}{!}{%
\begin{tabular}{ccccc}
\hline
\multirow{2}{*}{\textbf{Dependent Variable}} & \multicolumn{4}{c}{\textbf{Levene's Test Results}}                              \\ \cline{2-5} 
                                             & \textbf{F} & \textbf{df (Group)} & \textbf{df (Residual)} & \textit{\textbf{p}} \\ \hline
Trust in AI Confidence Level & 4.571 & 1 & 82 & 0.035* \\
Trust in AI Prediction       & 6.148 & 1 & 82 & 0.015* \\
Switch Percentage            & 2.005 & 1 & 82 & 0.161  \\
Under-reliance Percentage    & 0.206 & 1 & 82 & 0.651  \\
Over-reliance Percentage     & 0.405 & 1 & 82 & 0.526  \\
Accuracy Increase            & 0.004 & 1 & 82 & 0.949  \\ \hline
\end{tabular}
}
\end{table}

\begin{table}[h]
\caption{Assumption checks of the ANOVA about the main effects of information about AI confidence calibration in well-calibrated conditions. The significance levels are labeled ($p<0.05$: *, $p<0.01$: **, $p<0.001$: ***).}
\resizebox{.7\textwidth}{!}{%
\begin{tabular}{cllll}
\hline
\multirow{2}{*}{\textbf{Dependent Variable}} &
  \multicolumn{4}{c}{\textbf{Levene's Test Results}} \\ \cline{2-5} 
 &
  \multicolumn{1}{c}{\textbf{F}} &
  \multicolumn{1}{c}{\textbf{df (Group)}} &
  \multicolumn{1}{c}{\textbf{df (Residual)}} &
  \multicolumn{1}{c}{\textit{\textbf{p}}} \\ \hline
Trust in AI Confidence Level & 0.001 & 1 & 82 & 0.980  \\
Trust in AI Prediction       & 1.002 & 1 & 82 & 0.320  \\
Switch Percentage            & 1.815 & 1 & 82 & 0.182  \\
Under-reliance Percentage    & 4.698 & 1 & 82 & 0.033* \\
Over-reliance Percentage     & 0.259 & 1 & 82 & 0.612  \\
Accuracy Increase            & 0.032 & 1 & 82 & 0.858  \\ \hline
\end{tabular}
}
\end{table}

\begin{table}[h]
\caption{Assumption checks of the ANOVA about the main effects of information about AI confidence calibration in overconfident conditions. The significance levels are labeled ($p<0.05$: *, $p<0.01$: **, $p<0.001$: ***).}
\resizebox{.7\textwidth}{!}{%
\begin{tabular}{ccccc}
\hline
\multirow{2}{*}{\textbf{Dependent Variable}} & \multicolumn{4}{c}{\textbf{Levene's Test Results}}                              \\ \cline{2-5} 
                                             & \textbf{F} & \textbf{df (Group)} & \textbf{df (Residual)} & \textit{\textbf{p}} \\ \hline
Trust in AI Confidence Level & 2.320 & 1 & 82 & 0.132 \\
Trust in AI Prediction       & 1.830 & 1 & 82 & 0.180 \\
Switch Percentage            & 0.424 & 1 & 82 & 0.517 \\
Under-reliance Percentage    & 2.596 & 1 & 82 & 0.111 \\
Over-reliance Percentage                     & 15.190     & 1                   & 82                     & \textless{}0.001*** \\
Accuracy Increase            & 0.257 & 1 & 82 & 0.613 \\ \hline
\end{tabular}
}
\end{table}

\begin{table}[h]
\caption{One-way ANOVA results about the main effects of information about AI confidence calibration in underconfident conditions. The significance levels are labeled ($p<0.05$: *, $p<0.01$: **, $p<0.001$: ***).}
\resizebox{\textwidth}{!}{%
\begin{tabular}{ccccccc}
\hline
\multirow{2}{*}{\textbf{Dependent Variable}} &
  \multirow{2}{*}{\textbf{Statistical Test}} &
  \multicolumn{5}{c}{\textbf{Main Effect of Information about AI Confidence Calibration}} \\ \cline{3-7} 
 &
   &
  \multicolumn{1}{c}{\textbf{F}} &
  \multicolumn{1}{c}{\textbf{df (Group)}} &
  \multicolumn{1}{c}{\textbf{df (Residual)}} &
  \multicolumn{1}{c}{\textit{\textbf{p}}} &
  \multicolumn{1}{c}{\textbf{$\eta^2$}} \\ \hline
Trust in AI Confidence Level & Welch's ANOVA & 9.479 & 1 & 74.521 & 0.003** & 0.104 \\
Trust in AI Prediction       & Welch's ANOVA & 9.090 & 1 & 73.027 & 0.004** & 0.100 \\
Switch Percentage            & ANOVA         & 0.733 & 1 & 82     & 0.394   & 0.009 \\
Under-reliance Percentage    & ANOVA         & 1.741 & 1 & 82     & 0.191   & 0.021 \\
Over-reliance Percentage     & ANOVA         & 0.288 & 1 & 82     & 0.593   & 0.004 \\
Accuracy Increase            & ANOVA         & 0.078 & 1 & 82     & 0.780   & 0.001 \\ \hline
\end{tabular}
}
\end{table}

\begin{table}[h]
\caption{One-way ANOVA results about the main effects of information about AI confidence calibration in well-calibrated conditions. The significance levels are labeled ($p<0.05$: *, $p<0.01$: **, $p<0.001$: ***).}
\resizebox{\textwidth}{!}{%
\begin{tabular}{ccccccc}
\hline
\multirow{2}{*}{\textbf{Dependent Variable}} &
  \multirow{2}{*}{\textbf{Statistical Test}} &
  \multicolumn{5}{c}{\textbf{Main Effect of Information about AI Confidence Calibration}} \\ \cline{3-7} 
 &
   &
  \multicolumn{1}{c}{\textbf{F}} &
  \multicolumn{1}{c}{\textbf{df (Group)}} &
  \multicolumn{1}{c}{\textbf{df (Residual)}} &
  \multicolumn{1}{c}{\textit{\textbf{p}}} &
  \multicolumn{1}{c}{\textbf{$\eta^2$}} \\ \hline
Trust in AI Confidence Level & ANOVA         & 0.937 & 1 & 82     & 0.336 & 0.011 \\
Trust in AI Prediction       & ANOVA         & 0.946 & 1 & 82     & 0.334 & 0.011 \\
Switch Percentage            & ANOVA         & 0.178 & 1 & 82     & 0.674 & 0.002 \\
Under-reliance Percentage    & Welch's ANOVA & 0.845 & 1 & 74.623 & 0.361 & 0.010 \\
Over-reliance Percentage     & ANOVA         & 0.074 & 1 & 82     & 0.786 & 0.001 \\
Accuracy Increase            & ANOVA         & 0.133 & 1 & 82     & 0.716 & 0.002 \\ \hline
\end{tabular}
}
\end{table}

\begin{table}[h]
\caption{One-way ANOVA results about the main effects of information about AI confidence calibration in overconfident conditions. The significance levels are labeled ($p<0.05$: *, $p<0.01$: **, $p<0.001$: ***).}
\resizebox{\textwidth}{!}{%
\begin{tabular}{ccccccc}
\hline
\multirow{2}{*}{\textbf{Dependent Variable}} &
  \multirow{2}{*}{\textbf{Statistical Test}} &
  \multicolumn{5}{c}{\textbf{Main Effect of Information about AI Confidence Calibration}} \\ \cline{3-7} 
 &
   &
  \multicolumn{1}{c}{\textbf{F}} &
  \multicolumn{1}{c}{\textbf{df (Group)}} &
  \multicolumn{1}{c}{\textbf{df (Residual)}} &
  \multicolumn{1}{c}{\textit{\textbf{p}}} &
  \multicolumn{1}{c}{\textbf{$\eta^2$}} \\ \hline
Trust in AI Confidence Level & ANOVA         & 11.192 & 1 & 82     & 0.001**             & 0.120 \\
Trust in AI Prediction       & ANOVA         & 21.160 & 1 & 82     & \textless{}0.001*** & 0.205 \\
Switch Percentage            & ANOVA         & 28.192 & 1 & 82     & \textless{}0.001*** & 0.256 \\
Under-reliance Percentage    & ANOVA         & 9.515  & 1 & 82     & 0.003**             & 0.104 \\
Over-reliance Percentage     & Welch's ANOVA & 8.182  & 1 & 64.890 & 0.006**             & 0.091 \\
Accuracy Increase            & ANOVA         & 0.243  & 1 & 82     & 0.623               & 0.003 \\ \hline
\end{tabular}
}
\end{table}

\end{document}